\documentclass{article}

\usepackage{arxiv}

\usepackage[utf8]{inputenc} % allow utf-8 input
\usepackage[T1]{fontenc}    % use 8-bit T1 fonts
\usepackage{hyperref}       % hyperlinks
\usepackage{url}            % simple URL typesetting
\usepackage{booktabs}       % professional-quality tables
\usepackage{amsfonts}       % blackboard math symbols
\usepackage{nicefrac}       % compact symbols for 1/2, etc.
\usepackage{microtype}      % microtypography
\usepackage{lipsum}
\usepackage{authblk}
\usepackage[sort&compress,square,comma,numbers]{natbib}

\usepackage{enumitem}

\usepackage{float} 
\usepackage{times}
\usepackage{epsfig}
\usepackage{graphicx}
\usepackage{amsmath, amsfonts, amssymb}
\usepackage{bm}
\usepackage{algorithm}
\usepackage{algorithmicx}
\usepackage{comment}
\usepackage{fullpage}
\usepackage{tikz}
\usepackage{pifont}
\usepackage{amsthm}
\usepackage{subcaption}
\newcommand{\cmark}{\ding{51}}%
\newcommand{\xmark}{\ding{55}}

\usepackage{amsmath,amsfonts,amssymb}
\usepackage{mathtools}
\usepackage{multirow}
\usepackage[noend]{algpseudocode}
\usepackage{booktabs}
\usepackage[export]{adjustbox}
\usepackage{url}

\usepackage[textsize=scriptsize,backgroundcolor=green]{todonotes}
\newcommand\numberthis{\addtocounter{equation}{1}\tag{\theequation}}

\usepackage{xspace}
\newcommand*{\eg}{\textit{e.g.}\@\xspace}
\newcommand*{\ie}{\textit{i.e.}\@\xspace}

\newcommand*{\vs}{\textit{vs.}\@\xspace}
\newcommand*{\wrt}{\textit{w.r.t.}\@\xspace}
\makeatletter
\newcommand*{\etc}{%
	\@ifnextchar{.}%
	{\textit{etc}}%
	{\textit{etc.}\@\xspace}%
}
\makeatother
\algtext*{EndWhile}
\algtext*{EndFor}
\algtext*{EndIf}
\algdef{SE}[DOWHILE]{Do}{doWhile}{\algorithmicdo}[1]{\algorithmicwhile\ #1}%

\makeatletter
\def\BState{\State\hskip-\ALG@thistlm}
\makeatother

\title{Hybrid Neural Pareto Front (HNPF): A Two-Stage Neural-Filter Approach for Pareto Front Extraction}

\author[1]{\textbf{Gurpreet Singh} \textsuperscript{\dag}}
\author[2]{\textbf{Soumyajit Gupta} \textsuperscript{\dag}}
\author[3]{\textbf{Matthew Lease}}
\author[4]{\textbf{Clint Dawson}}
\affil[2]{Department of Computer Science}
\affil[3]{School of Information}
\affil[4]{Oden Institute for Computational Engineering and Sciences}
\affil[1]{The University of Texas at Austin}
\affil[ ]{\texttt{\{gurpreet, smjtgupta, ml\}@utexas.edu}, \texttt{clint.dawson@oden.utexas.edu}}

\begin{document}

\maketitle

{\let\thefootnote\relax\footnote{{\dag contributed equally to this work.}}}% under the supervision of \ddag.}}}

\begin{abstract}

Pareto solutions represent optimal frontiers for jointly optimizing multiple competing objective functions over the feasible set of solutions satisfying imposed constraints. Extracting a Pareto front is computationally challenging today with limited scalability and solution accuracy. Popular generic scalarization approaches do not always converge to a global optimum and can only return one solution point per run. Consequently, multiple runs of a scalarization problem are required to guarantee a Pareto front, where all instances must converge to their respective global optima. We propose a robust, low cost hybrid Pareto neural-filter (HNPF) optimization approach that is accurate and scales (compute space and time) with data dimensions, and the number of functions and constraints. A first-stage neural network first efficiently extracts a {\em weak} Pareto front, using Fritz-John conditions as the discriminator, with no assumptions of convexity on the objectives or constraints. A second-stage, low-cost Pareto filter then extracts the {\em strong} Pareto optimal subset from the {\em weak} front. Fritz-John conditions provide strong theoretical bounds on approximation error between the true and the network extracted {\em weak} Pareto front. Numerical experiments demonstrates the accuracy and efficiency of our approach.

\end{abstract}

\section{Introduction} \label{sec:intro}

%Decision making where multiple objectives must be simultaneously satisfied gives rise to 
Multi-Objective Optimization (MOO) problems arise frequently across diverse fields such as engineering \cite{marler2004survey}, finance \cite{tapia2007applications}, and supply chain management \cite{trisna2016multi}. 
%
% ML 2/6: further motivation not require and space better spent elsewhere.  If needed, move to subsequent section to tighten narrative here
%MOO is required for engineering decisions related to design choices in manufacturing, which lead to optimal throughput and quality control \cite{yang2012multi}. In the finance sector, these decisions lead to risk mitigation and return maximization for investment portfolios \cite{lakra2015multi,saborido2016evolutionary}. Balancing consumer supply and demand, or designing an optimal monetary policy design \cite{altiparmak2006genetic} for economic growth at different scales, also falls under this class of problems. 
Such problems share the common requirement to satisfy multiple competing objectives under a set of constraints imposed by physical or economic limits. 
%
% ML 2/6: redundant with next paragraph
%The difficulty in defining a suitable preference/selection criterion further gives rise to an optimal solution set instead of a unique solution.
A Pareto optimal solution \cite{pareto1906manuale} for an MOO problem is defined as the solution point away from which no single objective can be improved without diminishing at least one other objective. A Pareto front is then defined as the set of all such optimal points that satisfy this definition. Since all solutions reflect optimal tradeoff points between competing objective functions, choosing between solutions depends on the user's preferred tradeoff of objectives. %All elements of this solution set are equivalent, relegating a specific choice to the user preferences.

Computing a Pareto solution to an MOO problem requires optimizing competing (often non-convex) objective functions under constraints. This optimization problem is quite challenging: the solution set to an MOO can seldom be formulated as a closed-form expression, and solving this %. Additionally, generating this solution set 
for practical problems is compute intensive and often not feasible. Consequently, most research seeking practical solutions has focused on developing efficient approximations of the Pareto optimal solution set \cite{das1998normal,gobbi2015analytical,ghane2015new,pirouz2016computational}.

The ability to accurately and efficiently compute a Pareto optimal solution set, with theoretical guarantees and interpretability, would have tremendous practical value. For example, imagine an online news search for `influential CEOs' in which the user seeks results that are not only relevant, but also recent and reliable. In addition to optimizing for these three objective functions (information relevance, recency, and reliability), imagine the user further wishes to impose a demographic parity constraint on search results to ensure racial or gender parity  (since implicit data biases might otherwise yield search result coverage skewed toward white males). The Pareto optimal solution set would not only satisfy this parity constraint, but further enable the user to vary the composition of optimal search results based on their tradeoff preference between getting more recent, breaking news (which may be less reliable) \vs getting more reliable news (which may be less recent).

Our review of prior work reveals significant limitations: accuracy \cite{caton2020fairness}, compute time \cite{srinivas1994muiltiobjective,deb2002fast,miriam2020non}, and scalability, not to mention limited interpretability and verfiability. Recent methods for algorithmic fairness invoking Pareto optimality \cite{balashankar2019fair,lin2019pareto,martinez2020minimax,valdivia2020fair,wei2020fairness,xiao2017fairness} often suffer from inconsistent Pareto definitions and impractical assumptions of convex objective functions and constraints. Furthermore, a notable absence of benchmarks against known analytical forms makes it difficult to assess reported results, verify optimality, and A/B test alternative methods. This is in contrast to studies on computational methods \cite{das1998normal,ghane2015new,pirouz2016computational,gobbi2015analytical} in which such comparative benchmarking and verification is well established. Although accurate and verifiable, existing computational methods tend to generate Pareto points with low density (\ie providing a coarser representation of the underlying Pareto front) and poor scalability, with compute times ranging from hours to days as data dimensionality increases. 

% We draw inspiration from three seminal works: (1) \citet{das1998normal} proposed to break the functional domain boundary into uniform and evenly spaced segments (see CHIM in \cite{das1998normal}) to identify {\em weak} Pareto points with guarantees. Motivated by this, we first identify the {\em weak} Pareto front using a robust neural network (stage 1). (2) \citet{messac2003normalized} proposed the first Pareto filter to obtain the set of strong Pareto points from the aforementioned {\em weak} Pareto set. The filter uses an all-pair comparison criterion to reject dominated points from the {\em weak} Pareto set. This filter motivates our low-cost Pareto filter (stage 2) design which avoids the expensive all-pair comparison, using a plane search strategy. (3) \citet{gobbi2015analytical} presented the matrix form of the Fritz-John conditions satisfying the existence of Pareto points. Although their approach is only valid for convex cases, we extend the Fritz-John matrix form as a discriminator to identify Pareto point even for non-convex cases. 

In this work, we 
%first describe the problem formulation and survey prior work based on clear mathematical definitions for Pareto optimality. We then 
propose a novel two-stage architecture called Hybrid Neural Pareto Front (HPNF) for inducing Pareto optimal solution sets. Stage 1 consists of an interpretable and robust neural network that extracts a {\em weak} Pareto solution manifold as the output, given a dataset as input. Following this, Stage 2 provides a low-cost Pareto filter. For the network loss function, we use a discriminator based on Fritz-John conditions \cite{levi2006application} that accounts for multiple objectives and constraints. An approximate {\em weak} Pareto manifold is extracted as a weighted output of the \textit{softmax} function from the last layer of the network. The softmax activation classifies {\em weak} Pareto \vs non-Pareto data points. 

\begin{table}[bht]
    \centering
    % \resizebox{\columnwidth}{!}{%
    \begin{tabular}{c|cccc}
    \toprule
        \bf Method & \begin{tabular}[c]{@{}c@{}}\bf Generates only\\ \bf Pareto points\end{tabular} & \begin{tabular}[c]{@{}c@{}}\bf Generates \\\bf  Even Spread\end{tabular} & \begin{tabular}[c]{@{}c@{}}\bf Ease\\\bf  of Use\end{tabular} & \begin{tabular}[c]{@{}c@{}}\bf Efficient\\\bf  and Scalable\end{tabular} \\ \midrule
        Fair Pareto \cite{valdivia2020fair} & \xmark & \xmark & \xmark & \xmark \\
        mCHIM \cite{ghane2015new} & \cmark & \xmark & \xmark & \xmark \\
        PK \cite{pirouz2016computational} & \cmark &  \xmark & \xmark & \xmark \\
        NBI \cite{das1998normal} & \cmark & \cmark & \cmark & \xmark \\
        % Gobbi & \cmark & \cmark & \cmark & \cmark \\
        \hline
        \bf Our HNPF & \cmark & \cmark & \cmark & \cmark \\ \bottomrule
    \end{tabular}%}
    \caption{HNPF \vs existing state-of-the-art methods.}
    \label{tab:comp}
\end{table}

Our network architecture has few trainable parameters, making it robust to outliers and over-fitting.  Furthermore, we empirically show computational efficiency \vs current state-of-the-art methods \cite{das1998normal,gobbi2015analytical,pirouz2016computational}. Our approach produces only Pareto points (no false positives) with an even spread and higher density than possible with existing approaches. Furthermore, our approach is scalable with both increasing dimensions of the input data, and the number of functions and constraints. \textbf{Table \ref{tab:comp}} summarizes key properties of our HNPF approach \vs existing methods (see Section \ref{sec:related}). 

{\bf Contributions.} Our key contributions are as follows:
\begin{enumerate}[leftmargin=*]
    \item A manifold solution strategy for {\em weak} Pareto front identification based on Fritz-John conditions as the discriminator.
    \item A robust neural network for approximating the {\em weak} solution manifold for both convex and non-convex scenarios.
    \item Design of a computationally efficient Pareto filter to extract the {\em strong} Pareto set, compared to existing Pareto filters.
    \item Compared to other neural Pareto approaches our method extracts only Pareto optimal points with an even spread.
    \item HNPF is computationally scalable as the dimension of variable space, or functions and constraints increases.
    \item The final layer of the neural net is fully interpretable in terms of extracting the efficient set of input data as a manifold.
    \item The approximate {\em weak} Pareto is bounded below by $0 \leq \epsilon \leq 1$ \wrt the true manifold upon convergence.
    \item We will share our source code and benchmark datasets for reproducibility upon acceptance.
\end{enumerate}

%%%%%%%%%%%%%%%%%%%%%%%%%%%%%%%%%%%%%%%%%%%%%%%
%%%%%%%%%%%%%%%%%%%%%%%%%%%%%%%%%%%%%%%%%%%%%%%
%%%%%%%%%%%%%%%%%%%%%%%%%%%%%%%%%%%%%%%%%%%%%%%
\section{Related Work} \label{sec:related}

As noted earlier, since all Pareto optimal solutions reflect optimal tradeoff points between competing objective functions, choosing between solutions depends on the user's preferred tradeoff of objectives. Prior work can be organized around four directions for managing user preferences: 1) {\em No preference} \cite{zeleny1973compromise}: user preference criteria are not explicitly specified; 2) {\em a priori} \cite{gal1980multiple}: preference criteria are explicitly specified before computation; 3) {\em a posteriori} \cite{das1998normal}: preference criteria are explicitly specified after computation; and 4) {\em Interactive methods} \cite{miettinen2012nonlinear}: preference criteria are continuously consulted to isolate one of the optimal solutions. %For the example considered in Section \ref{sec:intro}, a user personalized choice might prioritize relevance over diversity, thereby selecting the $(90,10)\%$ pair over others although from a Pareto optimality point of view all choices are equivalent. \ml{read}

\subsection{Generic and Enhanced Scalarization}

One common approach is to convert an MOO problem into a Single Objective Optimization (SOO) problem via scalarization. However, generic scalarization methods \cite{balashankar2019fair,lin2019pareto,martinez2020minimax,valdivia2020fair,wei2020fairness} suffer from various limitations. Firstly, these approaches can only extract one solution point at a time given that the minimization problem converges to the global optimum. However for practical applications, with non-convex objectives and constraints, ensuring global optimality is non-trivial. Secondly, multiple runs with different trade-off parameters must be performed in order to extract the {\em weak} Pareto solution set, resulting in substantial computational overhead \cite{wei2020fairness}. Finally, the Pareto solution set can still form a non-convex manifold even when the objectives are convex \cite{ghane2015new} due to the presence of non-convex constraints (see Case III in Section \ref{sec:results}). These challenges prove to be major obstacles in the deployment of scalarization approaches as a practical tool for Pareto set extraction. 

Generic scalarization should not be confused with enhanced scalarization approaches \cite{das1998normal,ghane2015new,pirouz2016computational}, whose strength %does not lie particularly in the scalarization itself but rather 
lies in the specific localization of the objective space that allows treatment of non-convex functions and constraints. Although accurate and complete, enhanced scalarization approaches suffer from low computational scalability and low density of Pareto points on the solution manifold. For example, the $30$ dimensional benchmark in \textbf{Section \ref{sec:results} Case V} shows enhanced scalarization methods (mCHIM and PK) generating a Pareto set in approximately 18 hours.

Enhanced scalarization methods fall under category (3) of {\em a posteriori} methods. One such enhanced approach to solve an MOO involves constructing a local linear or epsilon scalarization based SOO. These methods include Normal Boundary Intersection (NBI) \cite{das1998normal}, Normal Constraint (NC) \cite{messac2003normalized}, Successive Pareto Optimization  \cite{mueller2009successive}, modified Convex Hull of Individual Minimum (mCHIM) \cite{ghane2015new} and Pirouz-Khorram (PK) \cite{pirouz2016computational}. NBI \cite{das1998normal} produces an evenly distributed set of Pareto points given an evenly distributed set of weights. Furthermore, NBI produces Pareto points in the non-convex parts of the Pareto curve while being independent of the relative scales of the objective functions. It uses the concept of Convex Hull of Individual Minima (CHIM) to break down the boundary/hull into evenly spaced segments and then trace the {\em weak} Pareto points. % on them. 

As an improvement over the NBI method, mCHIM uses a quasi-normal procedure to update the aforementioned CHIM set iteratively, to obtain a strong Pareto set. PK \cite{pirouz2016computational}, on the other hand, uses a local $\epsilon$-scalarization based strategy that searches for the Pareto front using controllable step-lengths in a restricted search region, thereby accounting for non-convexity. \citet{gobbi2015analytical} proposed a framework using Fritz-John conditions \cite{levi2006application} to obtain analytical solutions for convex functions and constraints with high point density. Note that, all of these aforementioned enhanced methods are guaranteed to converge to the Pareto front under their respective assumptions on the function property each method can handle.

\subsection{Bayesian and Genetic Approaches}

Methods that are {\em a priori} (2) require a prior distribution or initial seed parameters to be specified beforehand. Examples include Bayesian \cite{khan2002multi,calandra2014pareto,hernandez2016predictive} and Evolutionary  \cite{srinivas1994muiltiobjective,deb2002fast,miriam2020non} methods. \citet{khan2002multi}'s Bayesian method showed convergence to the Pareto front, but only under a linear setting, which is the strictest form of convexity. In recent Bayesian methods \cite{calandra2014pareto,hernandez2016predictive}, not only was convexity assumed, but even in actual convex cases significant error was still incurred. \citet{deb2002fast} introduced the Non-dominated Sorting Genetic Algorithm II (NSGA-II) algorithm that involves recombination, mutation and selection of a population representing the set of solutions points considered to be Pareto, each having one or more assigned objective values. The population is maintained to consist of diverse solutions, resulting in a set of non-dominated individuals that are expected to be near (not on) the real Pareto front. Other variants include NSGA-I \cite{srinivas1994muiltiobjective} and  NSGA-III \cite{miriam2020non}. However, convergence and reproducibility are not guaranteed with Genetic Algorithms, and significant hyper-parameter tuning is required.

\subsection{Approaches in the Fairness Literature}

%Recent work in fairness literature claim Pareto optimality but provide no proof of convexity, no benchmarking \vs analytical forms, and in some cases, seem to violate the definition of Pareto optimality itself.\ml{moved this paragraph here from earlier, don't know if we can cut} 

Pareto optimality is being increasingly pursued in classification and fairness research (\eg see survey \cite{caton2020fairness}). However, we are not aware of any work in this area providing verifiable solutions for benchmarking scenarios wherein the ground truth is known. 

Several works \cite{balashankar2019fair,martinez2020minimax} seek to balance classification accuracy \vs a \textit{no unnecessary harm} notion of fairness relying upon convexity assumptions without justification. The Weighted Sum Method (WSM) \cite{cohon2004multiobjective}, commonly used in fairness literature, is a linear scalarization approach to convert an MOO into an SOO using a convex combination of objective functions and constraints. However, this  is viable only when the functions and constraints are also convex \cite{ghane2015new}. 

\citet{valdivia2020fair} present a group-Fairness based trade-off model for decision tree based classifiers using the aforementioned genetic algorithm NSGA-II, which has the same convergence and reproducibility issues mentioned earlier. In addition, their reported results violate fundamental definitions of Pareto optimality. \citet{wei2020fairness} provide the first neural architecture for Pareto front computation for Fairness \vs Accuracy on classification datasets. They rely upon a Chebyshev scalarization, which assumes that objective functions must sum up to a constant, but do not justify this assumption. In the Fair-Recommendation literature,  \citet{xiao2017fairness} seek to balance \textit{social welfare} and {\em group fairness} for movie recommendations. They also propose a linear scalarization-based formulation which arrives at the true front for convex functions only. \citet{lin2019pareto} claim Pareto optimality using KKT conditions, which is guaranteed to converge only if the functions and constraints are convex under linear scalarization.

\subsection{Our Inspiration}

We draw inspiration from three seminal works: (1) \citet{das1998normal} proposed to break the functional domain boundary into uniform and evenly spaced segments (see CHIM in \cite{das1998normal}) to identify {\em weak} Pareto points with guarantees. Motivated by this, we first identify the {\em weak} Pareto front using a robust neural network (Stage 1). (2) \citet{messac2003normalized} proposed the first Pareto filter to obtain the set of strong Pareto points from the aforementioned {\em weak} Pareto set. The filter uses an all-pair comparison criterion to reject dominated points from the {\em weak} Pareto set. This filter motivates our low-cost Pareto filter (Stage 2) design, which avoids the expensive all-pair comparison using a plane search strategy. (3) \citet{gobbi2015analytical} presented the matrix form of the Fritz-John conditions satisfying the existence of Pareto points. Although their approach is only valid for convex cases, we extend the Fritz-John matrix form as a discriminator to identify Pareto point even for non-convex cases.

\section{Pareto Optimality} \label{sec:definition}

A general multi-objective optimization problem can formulated as:
\begin{align*}
    \underset{}{min} \quad F(x) &= (f_1(x),f_2(x),\ldots,f_k(x)) \numberthis \label{eq:multi}\\
    \text{s.t.} \quad x \in S &= \{ x \in \mathbb{R}^n | G(x)=(g_1(x), g_2(x),\ldots,g_m(x) \leq 0 \}
\end{align*}
in $n$ variables $(x_1,\ldots,x_n)$, $k$ objective functions $(f_1,\ldots,f_k)$, and $m$ constraint functions $(g_1,\ldots,g_m)$. Here, $S$ is the feasible set \ie the set of input values $x$ that satisfy the constraints $G(x)$. For a multi-objective optimization problem there is typically no single global solution, and it is often necessary to determine a set of points that all fit a predetermined definition for an optimum.

\subsection{Definitions}

We borrow and adapt existing definitions of Pareto optimality \cite{pareto1906manuale} from \citet{marler2004survey}.

\textbf{Strongly Pareto Optimal:} The Pareto optimal solution $x^*$ for Eq. \eqref{eq:multi} satisfies the  conditions:
\begin{align*}
    &\nexists x_j: f_p(x_j) \leq f_p(x^*), \quad \textrm{for} \quad p=1,2,\ldots,k\\
    &\exists l: f_l(x_j) < f_l(x^*) \numberthis \label{eq:pareto}
\end{align*}
This states that for $x^* \in X$ to be strongly Pareto optimal, there does not exist another $x_j \in X$, \textit{s.t.} $f_p(x_j) \leq f_p(x^*)$ for all functions $f_p, \forall p \in [1,k]$ and $f_l(x_j) < f_l(x^*)$ for at least one function $f_l$.

\textbf{Weakly Pareto Optimal:} A {\em weak} Pareto point $\tilde{x}^*$ satisfies:
\begin{align}
    \nexists x_j: f_p(x_j) < f_p(\tilde{x}^*), \quad \textrm{for} \quad p=1,2,\ldots,k
\end{align}
A point is weakly Pareto optimal if no other point exists that improves all of the objectives simultaneously. This is different from a strongly Pareto optimal point, \textit{s.t.} no point exists, that improves at least one objective without detriment to other objectives. 

\textbf{Efficient and Inefficient Points:} A Pareto efficient point is defined on the domain $X$ as a point $x^{*}$ \textit{iff} no other point $x$ exists \textit{s.t.} $F(x)\leq F(x^*)$ with at least one $f_{p}(x) < f_{p}(x^{*})$. The point $x^*$ is considered inefficient otherwise. 

\textbf{Dominated and Non-Dominated Points:} Dominated points are defined \wrt the objective function in the criterion space. The objective function vector $F(x^{*})$ is non-dominated \textit{iff} no other vector $F(x)$ exists \textit{s.t.} $F(x)\leq F(x^{*})$ with at least one $f_{p}(x)<f_{p}(x^*)$. The vector $F(x^*)$ is considered dominated otherwise. 

\begin{figure}[ht]
    \centering
    \begin{subfigure}{.35\linewidth}
      \centering
      \includegraphics[width=\linewidth]{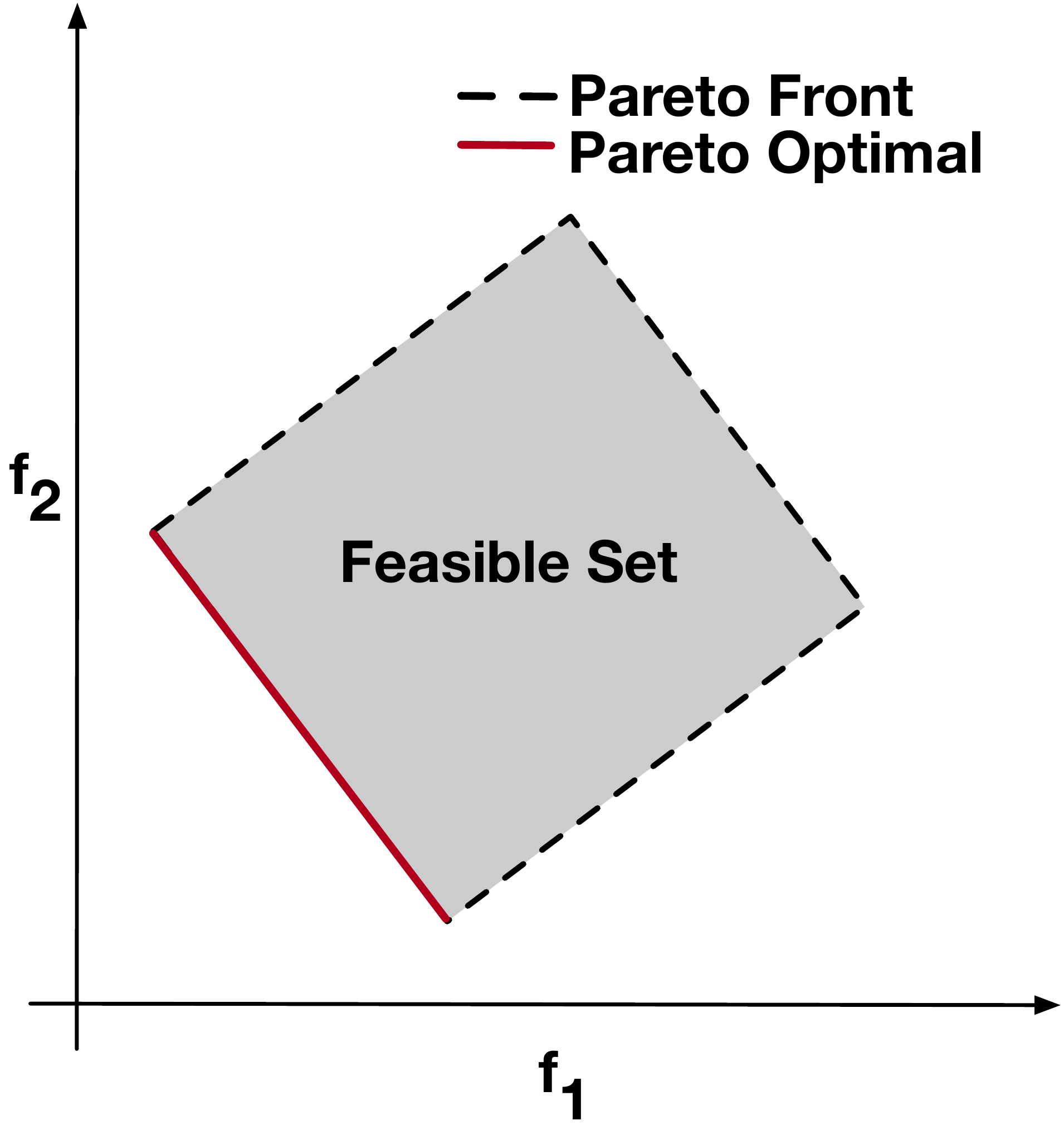}
      \caption{Convex form}
    \end{subfigure} \qquad 
    \begin{subfigure}{.35\linewidth}
      \centering
      \includegraphics[width=\linewidth]{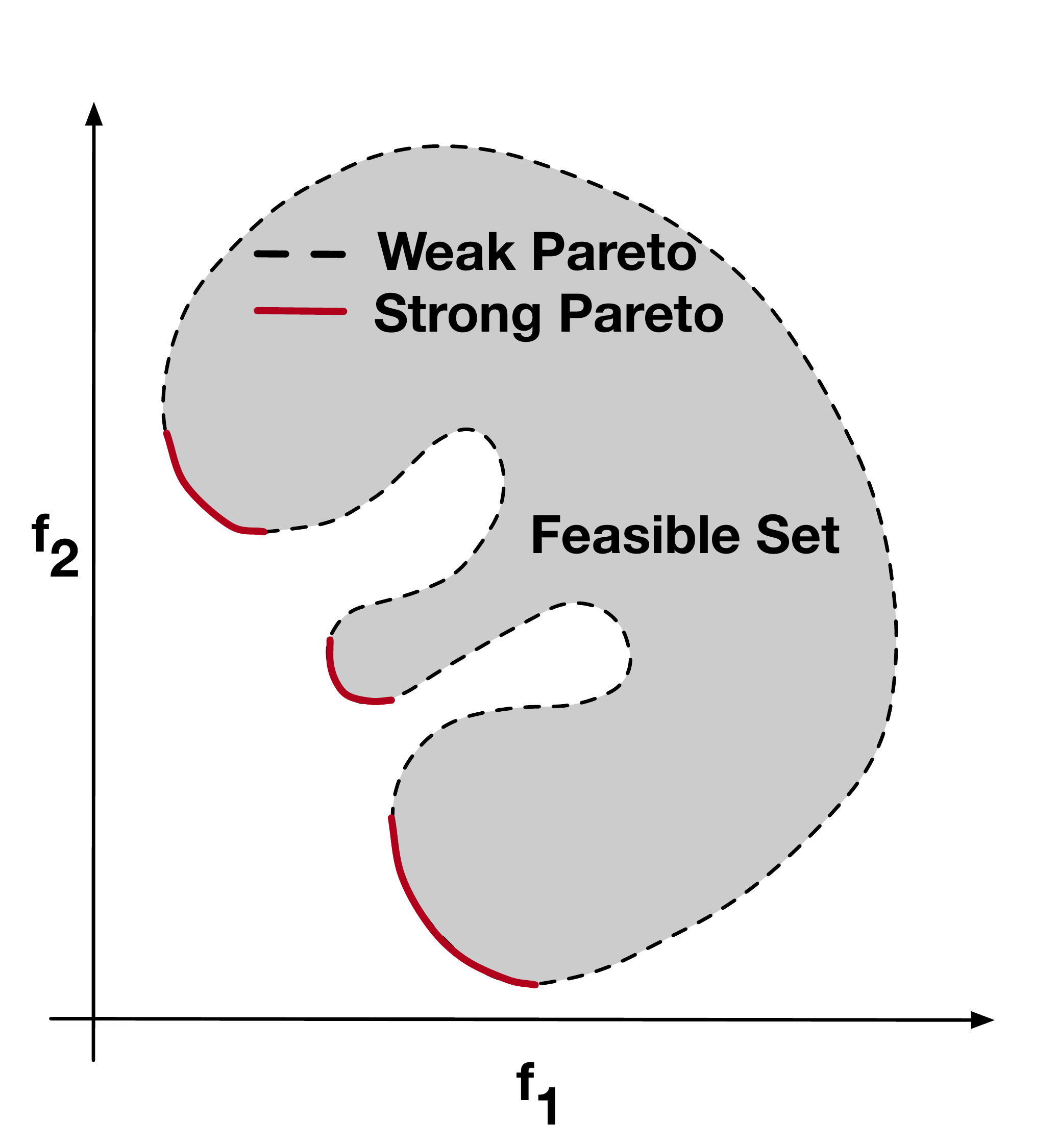}
      \caption{Non-Convex form}
    \end{subfigure}
    \caption{Pareto optimal set under different objectives. Note that the red line corresponds to the min-min strong Pareto optimal front for both problems and the dashed line corresponds to the {\em weak} Pareto Front for the entire Feasible Set.}
    \label{fig:par_set}
\end{figure}

\textbf{Fig. \ref{fig:par_set}} shows two separate MOOs for two competing functions. The feasible set of points are shown as gray shaded area with the Pareto boundary shown as dashed lines. For both problems, a joint minimization problem is considered, resulting in a Pareto optimal set (red curves) facing the origin. A different Pareto boundary can be obtained if a joint-maximization or a mixed min-max problem is considered. As shown in Fig. \ref{fig:par_set} (b), a strong Pareto optimal solution from a non-convex {\em weak} Pareto front can result in a discontinuous manifold. In the following sections, although we rely on dominated and non-dominated points for visualization purposes, the approximation errors and our algorithm rely primarily on efficient and inefficient points for computational purposes.

\subsection{Fritz John Conditions}

Let the objective and constraint function in Eq. \eqref{eq:multi} be differentiable once at a decision vector $\tilde{x}^* \in \mathcal{S}$. The Fritz-John necessary conditions for $\tilde{x}^*$ to be {\em weak} Pareto optimal is that vectors must exists for $0 \leq \lambda \in \mathbb{R}^k$, $0 \leq \mu \in \mathbb{R}^m$ and $(\lambda, \mu) \neq (0,0)$ (not identically zero) \textit{s.t.} the following holds:
\begin{align*}
    \sum_{i=1}^k \lambda_i \nabla f_i(\tilde{x}^*) + \sum_{j=1}^m \mu_j \nabla g_j(\tilde{x}^*) = 0 \numberthis \label{eq:fjcond} \\
    \mu_jg_j(\tilde{x}^*) = 0, \forall j=1,\ldots,m
\end{align*}
\citet{gobbi2015analytical} presented an $L$ matrix form, comprising the gradients of the functions and constraints as follows: \begin{align*}
&L = \begin{bmatrix}
\nabla F & \nabla G \\
\mathbf{0} & G
\end{bmatrix} \quad [(n+m) \times (k+m)] \label{eq:fjmat} \numberthis \\
&\nabla F_{n \times k} = [\nabla f_1, \ldots, \nabla f_k]\\
&\nabla G_{n \times m} = [\nabla g_1, \ldots, \nabla g_m]\\
&G_{m \times m}=diag(g_1,\ldots,g_m) 
\end{align*}
The matrix equivalent of Fritz John Conditions for $x^*$ to be Pareto optimal, is to show the existence of $\lambda \in \mathbb{R}^{k+m}$ in Eq. \eqref{eq:fjcond} such that:
\begin{align}
    L \cdot \delta = 0 \quad \text{s.t.} \quad L=L(\tilde{x}^*),\mathbf{\delta} \geq 0, \mathbf{\delta} \neq 0 \label{eq:fjmatrix}
\end{align}
The non-trivial solution ($\mathbf{\delta}$ is not identically zero) for Eq. \eqref{eq:fjmatrix} is:
\begin{align}
    det(L^TL)=0 \label{eq:paropt}
\end{align}
The {\em weak} Pareto front is characterized by the set of points such that matrix $L$ is low rank. This ensures that the points identified are either inside the feasible set or at boundaries dictated by the constraints. For \eg if $\mu_{1} = 0$ for any $\lambda_{i}$, then $\sum_{i} \lambda_{i} f_{i} = 0$ must be satisfied for the corresponding internal point $x^{*}$ to be Pareto. Similarly if $\mu_{1} \neq 0,\mu_{j\neq 1}=0$ in the aforementioned case, then $g_{1} = 0$ holds true for the corresponding boundary point $x^{*}$ to be Pareto. Note that all Pareto points satisfy $\nabla f_{i} = 0$ for at least one $i$ whether they lie inside the feasible set or on the boundaries. This is to say that all points $x^{*}$ are local optimizers for at least one $f_{i}$. The rank of the matrix $L^{T}L$ determines the dimension of the Pareto manifold. Furthermore, the necessary condition written as $det(L^{T}L)$ is independent of the preference parameters $\lambda_{i}, \, \mu_{j}$. Eq. \eqref{eq:paropt} now serves as a condensed discriminator to identify a {\em weak} Pareto front. In what follows, we use this matrix form of Fritz-John conditions to approximate the Pareto front using a robust, low-weight, neural network.

\section{HNPF Framework}

% In this section we lay out the details of the proposed two-stage architecture for Pareto set detection. Both stages are computationally efficient compared to existing methods and applies to all forms (convex and non-convex) of functions and constraints.
In this section we lay out the details of the proposed computationally efficient hybrid two-stage architecture for Pareto set detection.

\subsection{Stage 1: Neural Net for Weak Pareto Front}
\label{sec:stage1}

\begin{figure}[ht]
    \centering
    \includegraphics[width=0.6\linewidth]{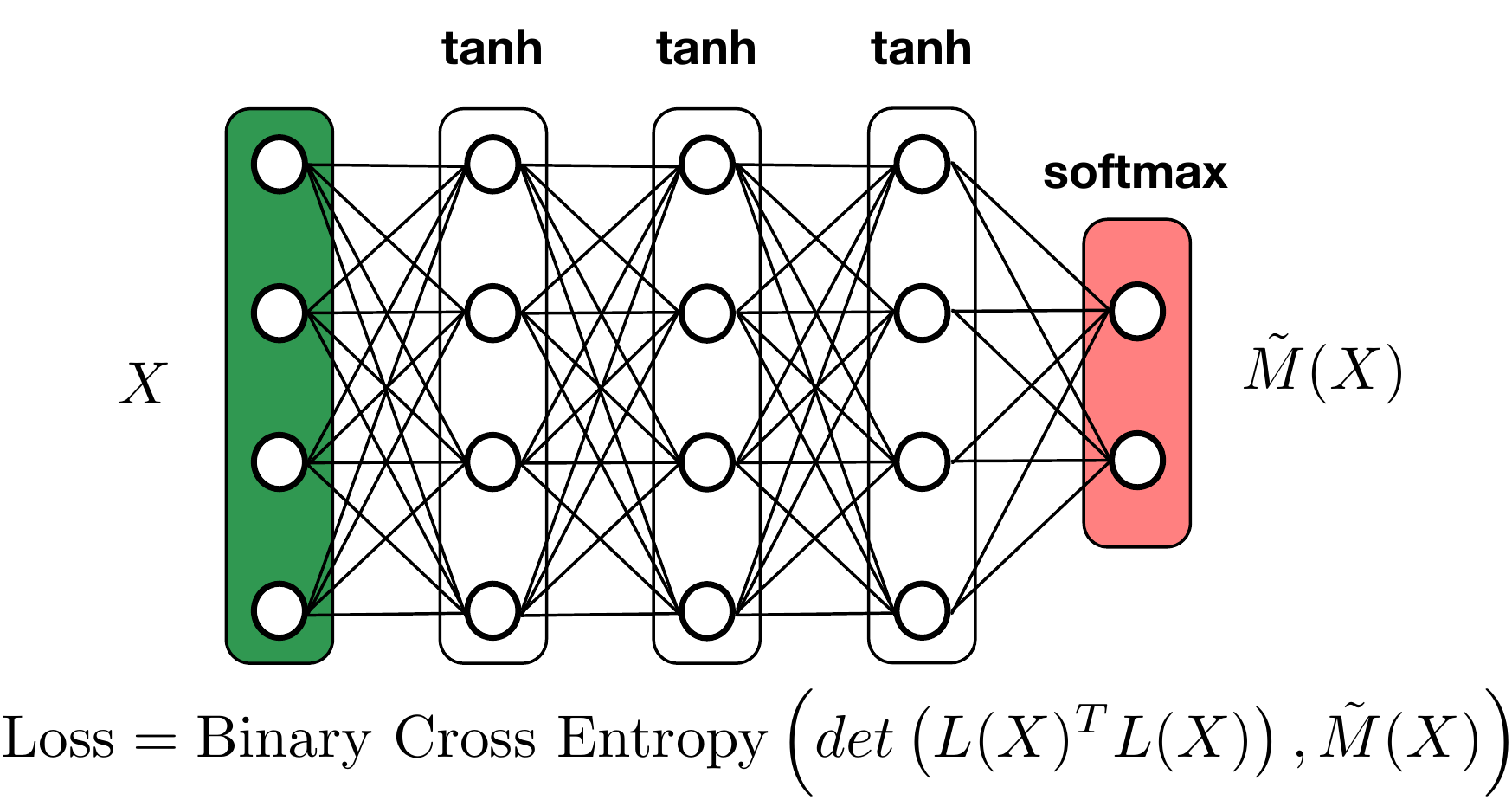}
    \caption{Proposed Neural network for finding the {\em weak} Pareto points. The last layer performs Fritz John criteria enabled binary cross entropy classification of data points, using softmax activation, ensuring {\em weak} Pareto optimality.}
    \label{fig:arch}
\end{figure}

The proposed neural network consists of feed-forward layers with \textit{tanh} activation. There are three layers of dense connections with eight neurons each, to smoothly approximate the optimal solution manifold $M(X)$ as $\tilde{M}(X)$, shown in Fig. \ref{fig:arch}. The last layer of the network has two neurons with \textit{softmax} activation for binary classification of Pareto \vs non-Pareto points. In other words, this layer approximates the separation manifold that distinguishes inefficient points from the {\em weak} Pareto points, in the feasible set $S$. Note that our network loss is representation driven, since the Fritz John discriminator (Eq. \eqref{eq:paropt}) explicitly classifies points as being {\em weak} Pareto or not. The network accepts or rejects the input data points $X$ based on the Fritz-John discriminator described by the objective functions and constraints. The Fritz-John necessary conditions for {\em weak} Pareto optimality, as pointed out earlier, require that the $D=det(L(X)^{T}L(X))=0$. Therefore, $1-D$ and $D$ naturally provide us with binary labels for the \textit{softmax} activated output layer. A binary cross entropy loss ensures that the distribution of the extracted manifold $\tilde{M}(X)$ matches the distribution of the {\em weak} Pareto front satisfying the Fritz-John conditions. The network architecture is purposely kept low weight for {\em weak} Pareto manifold extraction to provide robustness against outliers.

\textbf{Error bound.} For a user-prescribed relaxation margin $0 \leq \epsilon \leq 1$, the approximation error between the network extracted manifold $\tilde{M}(\tilde{X})$ and the true solution $M(X^*)$ is bounded below by  $\|\tilde{M}(\tilde{X}) - M(X^*)\|_2 \leq \epsilon$, upon convergence. See \textbf{Appendix \ref{app:error}} for proof.

\subsection{Stage 2: Pareto filter for Strong Pareto Set}
\label{sec:filter}

A Pareto filter is an algorithm that, given a set of {\em weak} Pareto points in objective space, retains a subset of non-dominated points. This corresponds to the strong Pareto set \textit{s.t.} none of the points are dominated. In other words, the filter eliminates all dominated points from the given set. A state of the art Pareto filter, defined in \citet{messac2003normalized}, is used as a post-processing step to extract a strong Pareto set. However, note that this Pareto filter requires an all-pairs comparison, an $O(n!)$ calculation, and thus becomes computationally expensive as the point set grows. This necessitates that the number of {\em weak} Pareto points be small when using this filter. However, since Stage 1 of our approach generates {\em weak} Pareto points with high density, the filter proves to be quite expensive.

We present an efficient Pareto filter algorithm for finding a strong Pareto set which is computationally scalable to arbitrary dimensions. The algorithm is based on a plane search strategy, inspired by Kd-Trees \cite{bentley1990k}, well known for efficient data partitioning and storage. The compute complexity of our approach is determined by the number of competing functions while being linearly proportional to the number of points. The inputs to the algorithm (\textbf{Alg. \ref{alg:filter}}) are the number of functions $k$, their minimum and maximum bounds $f(min),f(max)$, discretization level $h$ of the function space and the {\em weak} Pareto points $P$. The output is the strong Pareto set $\{x^*\}$.

Following Alg. \ref{alg:filter}, we can estimate its time complexity. There are three nested \textit{for} loops which carry the load. In the worst case that the points in the {\em weak} Pareto set are all strong Pareto, then the cardinality of the set $P$ remains unchanged. Let us also define $z = (f_i(max)-f_i(min))/h$ to denote the number of chunks into which the function space is divided. Thus, the worst case complexity of the proposed Pareto filter is $\mathbf{O(kzn)}$. For scenarios, where the strong Pareto set is a subset of the original {\em weak} set $P$, the complexity reduces in the factor guided by $n$.

\begin{algorithm}[htb]
	\caption{Pareto filter}
% 	\algsetup{linenosize=\tiny}
    \footnotesize
	\begin{algorithmic}[1]
	\BState \textbf{Data} $P=\{\tilde{x}^*\} \in \mathbb{R}^{n}$ weak Pareto points
	\BState \textbf{Input} $f(min),f(max):$ bounds of each function $f_i,\forall i \in k$
	\BState \textbf{Input} $k:$ number of functions, $h:$ discretization level
	\For {$i \in k$} \Comment{Loop over all functions}
	    \State $level=f_i(min)$
	    \For {$j \in (f_i(max)-f_i(min))/h)$} \Comment{Loop over all levels}
	        \State $temp=\varnothing$
	        \For {$p \in P$}
        	    \If {$level \leq f_i(p) < level+h$}
        	        \State $temp = temp \cup p$
        	    \EndIf
        	\EndFor
    	    \If {$card(temp) > 1$} \Comment{Cardinality of set}
    	        \State $x_p = min \, f_q(x), x \in temp,q=i+1$ \Comment{Efficient point}
    	        \State $P = P \backslash (temp \backslash x_p)$ \Comment{Remove inefficient points}
    	   % \ElsIf {$card(temp) == 1$}
    	   %     \State $interim = interim \cup temp$
    	    \EndIf
    	    \State $level = level+h$
    	 \EndFor
	\EndFor
	\BState \textbf{Output}: Strong Pareto set $x^*=P$ 
	\end{algorithmic} \label{alg:filter}
%    \vspace{-1em}
\end{algorithm}

\begin{figure}[ht]
    \centering
    \begin{subfigure}{0.32\linewidth}
      \centering
      \includegraphics[width=\linewidth]{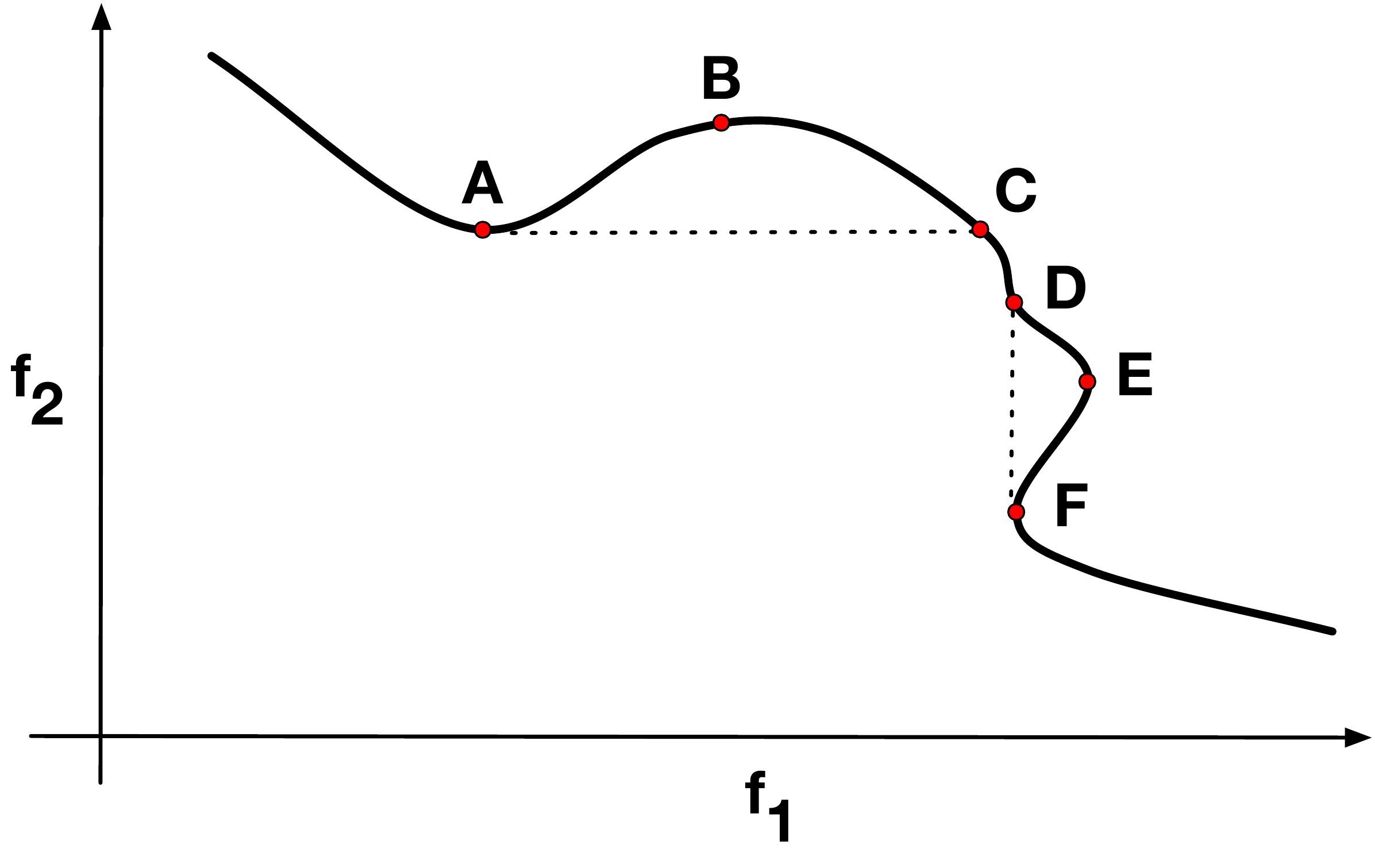}
      \caption{Weak Pareto Set}
    \end{subfigure}
    \begin{subfigure}{0.32\linewidth}
      \centering
      \includegraphics[width=\linewidth]{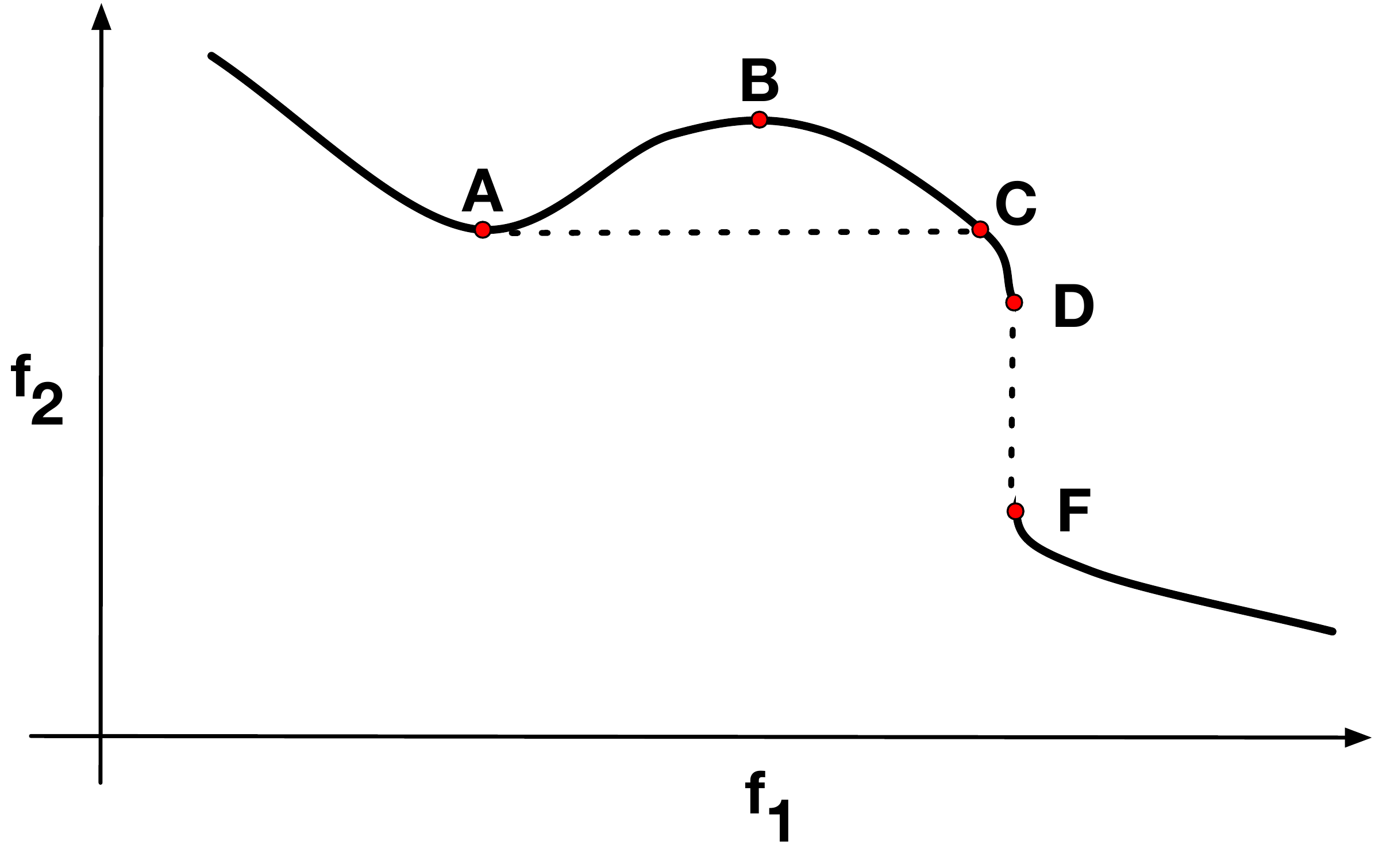}  
      \caption{First pass}
    \end{subfigure}
    \begin{subfigure}{0.32\linewidth}
      \centering
      \includegraphics[width=\linewidth]{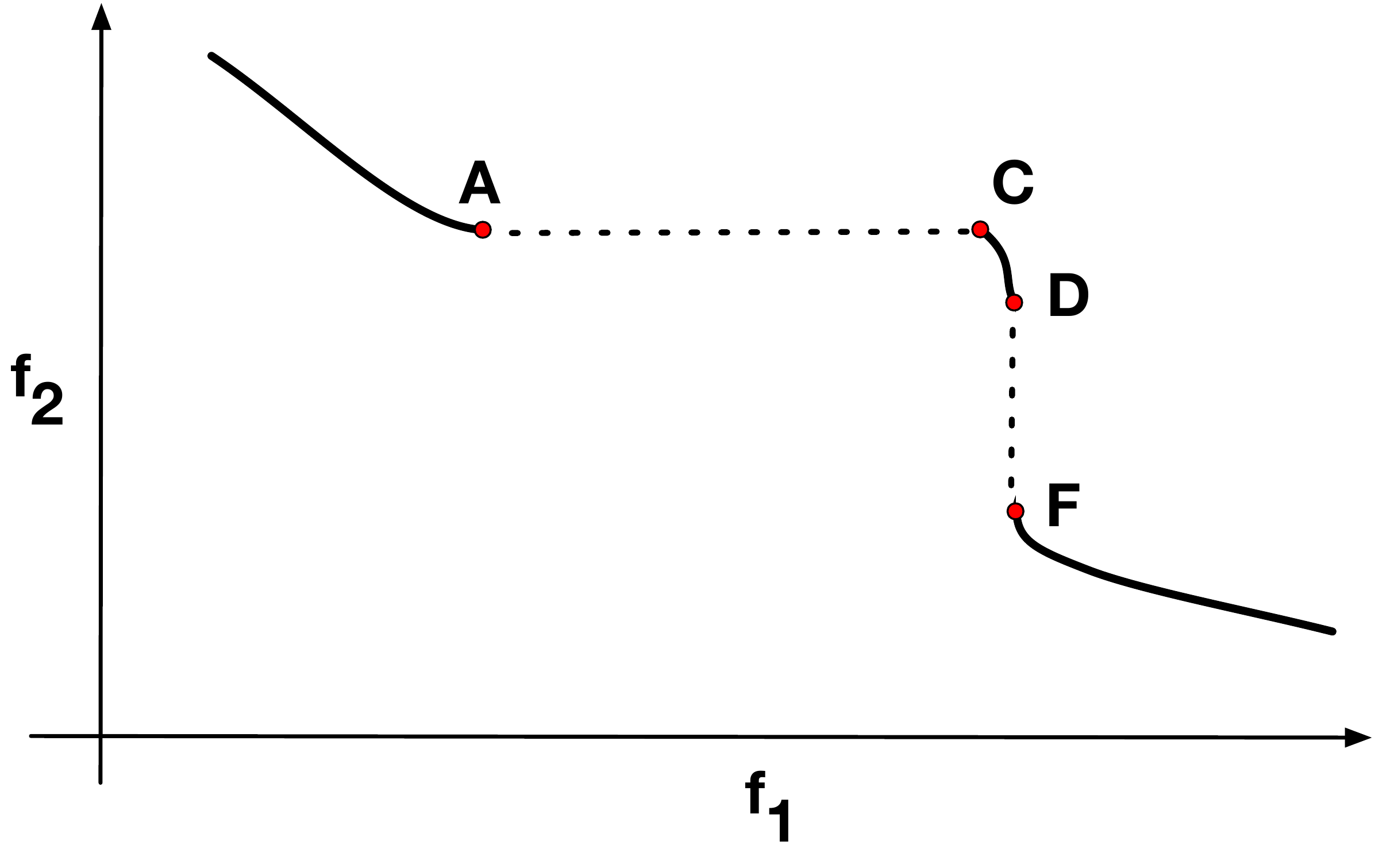}  
      \caption{Second pass}
    \end{subfigure}
    \caption{Illustration of the proposed Pareto Filter on a two-function scenario for visualization purposes}
    \label{fig:par_filter}
\end{figure}

It is easy to visualize the working of the proposed Pareto filter in \textbf{Fig. \ref{fig:par_filter}}. We start with the {\em weak} Pareto set $P$ (Fig. \ref{fig:par_filter} (a)) for a non-convex form. The first pass over $f_1$ removes a set (segment DEF) of dominated Pareto points (Fig. \ref{fig:par_filter} (b)). The leftover points in $P$ are then filtered again based on $f_2$ (Fig. \ref{fig:par_filter} (c)), where the dominated points (segment ABC) as per $f_2$ are removed. Points surviving the filtering process belong to the strong Pareto set. 

%=========================================
%=========================================

%%%%%%%%%%%%%%%%%%%%%%%%%%%%%%%%%%%%%%%%%%
%%%%%%%%%%%%%%%%%%%%%%%%%%%%%%%%%%%%%%%%%%
%%%%%%%%%%%%%%%%%%%%%%%%%%%%%%%%%%%%%%%%%%
\section{Results}
\label{sec:results}

In this section, we present five numerical experiments for benchmarking and analysis (see \textbf{Appendix \ref{app:cases}} for two additional cases). These experiments address standard analytical forms, with increasing complexity and scale in the number of \textit{functions} $(k)$, \textit{constraints} $(m)$ and \textit{dimension} of variables $(n)$. While cases may appear synthetic, they arise from practical physical domains in various engineering fields. We compare our results \vs those from two current state-of-the-art methods: mCHIM \cite{ghane2015new} and PK \cite{pirouz2016computational}.

%\subsection{Sampling}
\textbf{Sampling.} Since we are only provided with objective functions and constraints, we must sample data points from the variable domain in order to generate candidates to test for optimality. Firstly, if there are any direct constraints on variable values, we consider this feasible domain for sampling, as in the benchmark cases. Secondly, lacking any prior knowledge of where the Pareto front may reside, we sample values random uniform distribution in the feasible variable domain. %Thirdly, candidates are tested against constraint functions, and any any non-feasible values are excluded without ever invoking objective functions. Finally, remaining feasible 
Objective functions evaluated at these points generate a quantized, topographic map of the function domain that is then used to identify optimal points. For each benchmark test case below, we generate 11K points from a random uniform distribution in the feasible variable domain, to serve as training data. %Non-feasible values (violating constraints) are then detected and filtered out\ml{let's discuss}. 
The training-validation split is 90-10\%. Once the manifold is learned by the network, we feed in 90k points within the permissible domain to plot the Pareto set for visualization. 

\subsection{Experimental Setup}

Experiments use an Nvidia 2060 RTX Super 8GB GPU, Intel Core i7-9700F 3.0GHz 8-core CPU and 16GB DDR4 memory. We use the Keras \cite{chollet2015} library on a Tensorflow 2.0 backend with Python 3.7 to train the networks in this paper. For optimization, we use AdaMax \cite{kingma2014adam} with parameters (\textit{lr}=0.001) and $1000$ steps per epoch. 

%The training and testing datasets consists of 10k and 90k points respectively, for all cases, with points drawn from a random distribution. The training-validation split was set to 90-10\%. %Under this setting we roughly recover 500 weak Pareto points on test data. 

While neural approaches often pre-initialize the network with layer-wise training \cite{bengio2007greedy}, a strength of HPNF is that all network weights can be simply drawn from a uniform random distribution. Since the data domain is discrete, an exact zero might not be achievable. We therefore use a slightly relaxed criterion of $\epsilon=0.001$ as the classification margin. Any point above this value will be classified as {\em weak} Pareto. For all  results, the extracted Pareto set (shaded red) overlaps the true Pareto set with an $\epsilon$ spread.

Due to stochastic variation, neural network studies often report variance across several runs. However, the only approximation errors with our method lie in the extracted manifold over runs. Error bounds are given in Section \ref{sec:stage1} and Appendix \ref{app:error}. Since the manifold remains constant across runs, the loss itself is the approximation error with a minimum achievable value of 0 at machine precision. With respect to this 0 over multiple runs, the loss function is the deviation from the true manifold. We thus do not report mean-variance across runs. Section \ref{sec:profile} shows loss profiles.

\subsection{Case I: n=2, k=2, m=2}

\setlength\abovedisplayskip{0.25em}
\setlength\belowdisplayskip{0.25em}

\begin{figure}[ht]
    \centering
     \begin{subfigure}{0.32\linewidth}
      \centering
      \includegraphics[width=\linewidth]{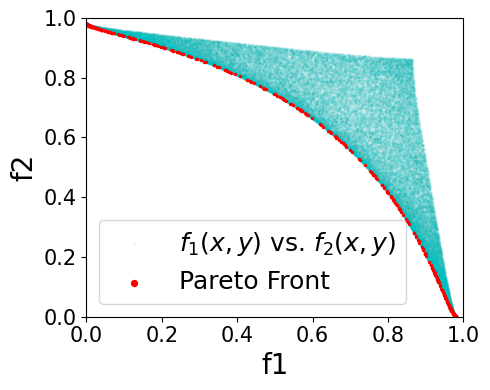}
      \caption{Function Domain}
    \end{subfigure} \qquad
    \begin{subfigure}{0.32\linewidth}
      \centering
      \includegraphics[width=\linewidth]{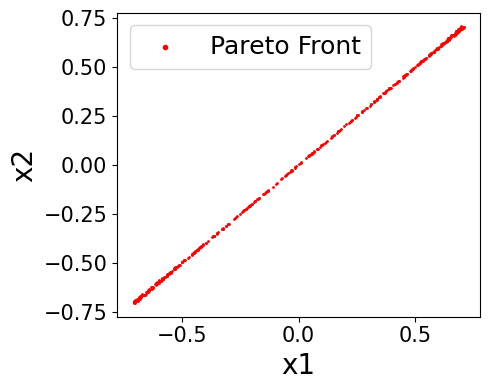}
      \caption{Variable Domain}
    \end{subfigure}
    \caption{Pareto Front for Case I. Please refer to color plots for proper visualization for all following figures.}
    \label{fig:pareto1}
\end{figure}

This problem was originally proposed in \cite{fonseca1998multiobjective}. Jointly minimize
\begin{align*}
    &f_1(x_1,x_2) = 1 - exp(-[(x_1-1/\sqrt(2))^2 + (x_2-1/\sqrt(2))^2]) \\
    &f_2(x_1,x_2) = 1 - exp(-[(x_1+1/\sqrt(2))^2 + (x_2+1/\sqrt(2))^2]) \\
    &\text{s.t.} \quad g_1,g_2: -1/\sqrt{2} \leq x_1,x_2 \leq 1/\sqrt{2}
\end{align*}
% The analytical solution of the Pareto front is:
% \begin{align*}
%     f_1 = 1 + (f_2-1) \exp(-4+4 \sqrt{- \log (1-f_2)}) \\
%     \text{under} \quad 0 \leq f_2 \leq 0.982, -\frac{1}{\sqrt{2}} \leq x_1,x_2 \leq \frac{1}{\sqrt{2}}
% \end{align*}
The Pareto set can be computed by applying linear scalarization (WSM) for this problem since all the functions and constraints are convex. \citet{gobbi2015analytical}, NBI, mCHIM and PK are able to extract the Pareto solution set. \textbf{Fig. \ref{fig:pareto1}} shows the solution from our proposed network with high point density, where we can visually verify (Fig. \ref{fig:pareto1}(b)) that the network approximated the Pareto manifold, validating that the Pareto points indeed closely satisfy $x_1=x_2$.

\subsection{Case II: n=2, k=2, m=2}

\begin{figure}[ht]
    \centering
     \begin{subfigure}{0.32\linewidth}
      \centering
      \includegraphics[width=\linewidth]{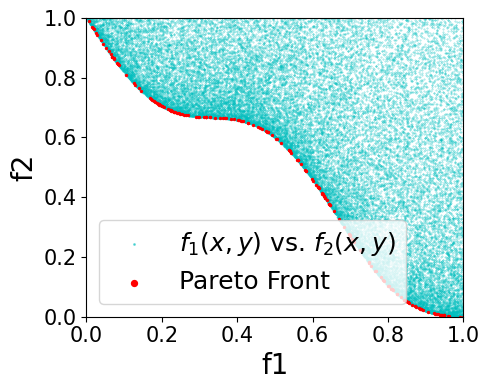}
      \caption{Function Domain}
    \end{subfigure} \qquad
    \begin{subfigure}{0.32\linewidth}
      \centering
      \includegraphics[width=\linewidth]{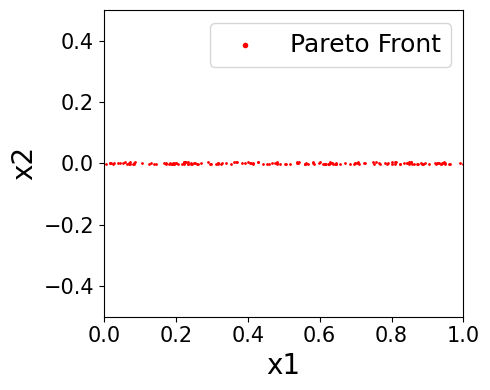}
      \caption{Variable Domain}
    \end{subfigure} \\
    \begin{subfigure}{0.32\linewidth}
      \centering
      \includegraphics[width=0.78\linewidth]{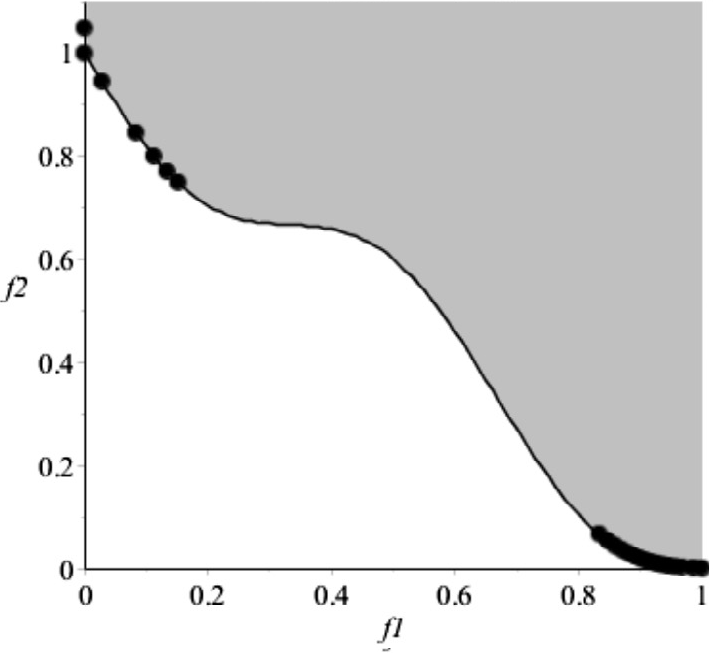}
      \caption{WSM}
    \end{subfigure}
    \begin{subfigure}{0.32\linewidth}
      \centering
      \includegraphics[width=0.8\linewidth]{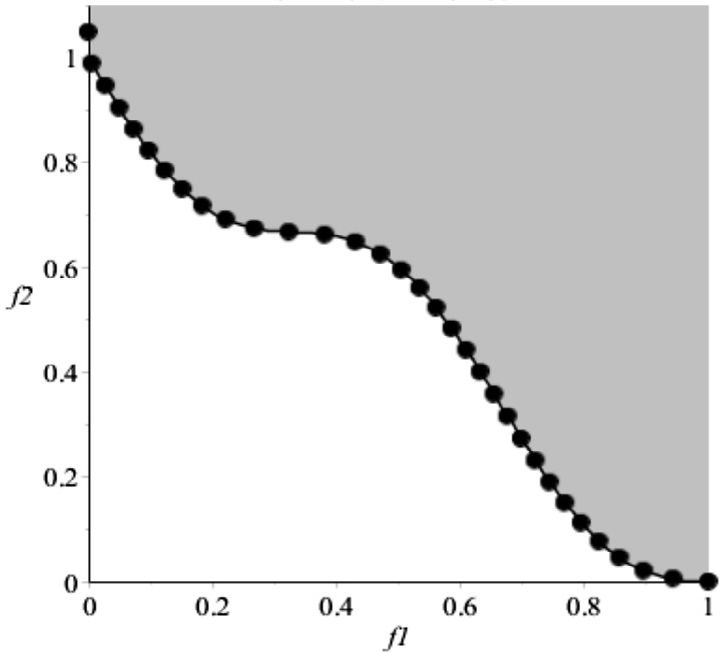}
      \caption{NBI}
    \end{subfigure}
    \begin{subfigure}{0.32\linewidth}
      \centering
      \includegraphics[width=0.76\linewidth]{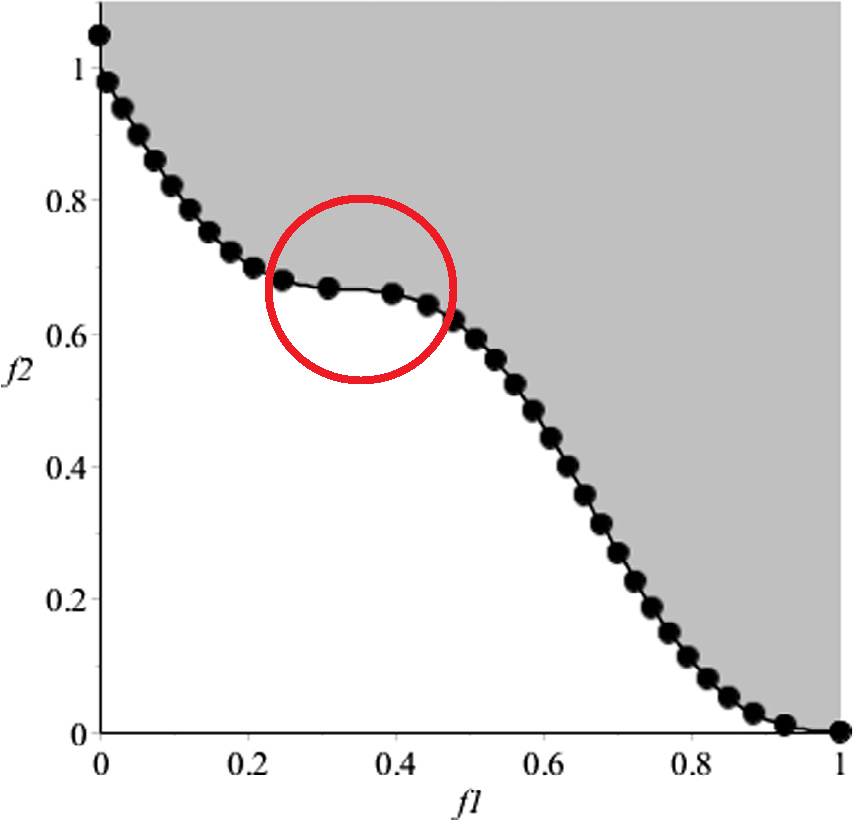}
     \caption{mCHIM}
    \end{subfigure}
    \caption{Pareto Front for Case II. Note the even spread of point HNPF produces. NBI produce an even spread while mCHIM cannot. WSM fails for non-convexity in function.}
    \label{fig:pareto2}
\end{figure}

This problem was proposed in \cite{ghane2015new}. Jointly minimize
%\vspace{-1em}
\begin{align*}
    &f_1(x_1,x_2) = x_1 \\
    &f_2(x_1,x_2) = 1 + x_2^2 - x_1 - 0.1sin 3 \pi x_1\\
    &\text{s.t.} \quad g_1,g_2: 0 \leq x_1 \leq 1, -2 \leq x_2 \leq 2
\end{align*}
\citet{gobbi2015analytical} does not consider this scenario due to non-convexity of $f_2$. As shown in \cite{ghane2015new}, WSM can only identify a subset of the Pareto optimal points. NBI, mCHIM and PK methods are able to identify points in this case with equal density. \textbf{Fig. \ref{fig:pareto2}} shows the results from our model with high point density. It also satisfies closely the true Pareto manifold given by $0 \leq x_1 \leq 1,x_2=0$ in Fig. \ref{fig:pareto2}(b).

\subsection{Case III: n=2, k=2, m=4}

\begin{figure}[ht]
    \centering
     \begin{subfigure}{0.32\linewidth}
      \centering
      \includegraphics[width=\linewidth]{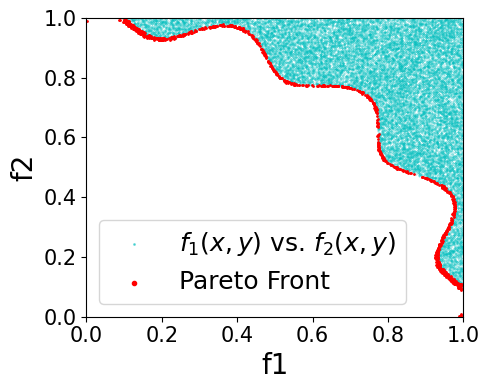}
      \caption{Dominated}
    \end{subfigure}
    \begin{subfigure}{0.32\linewidth}
      \centering
      \includegraphics[width=\linewidth]{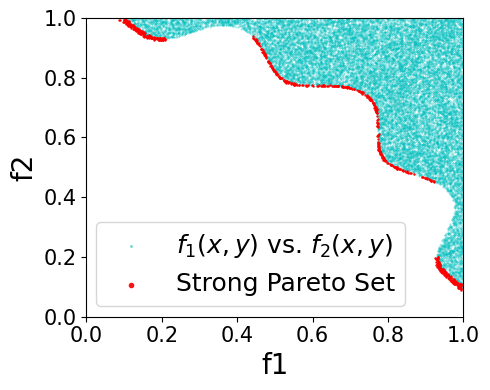}
      \caption{Non-Dominated}
    \end{subfigure}
    \begin{subfigure}{0.32\linewidth}
      \centering
      \includegraphics[width=0.77\linewidth]{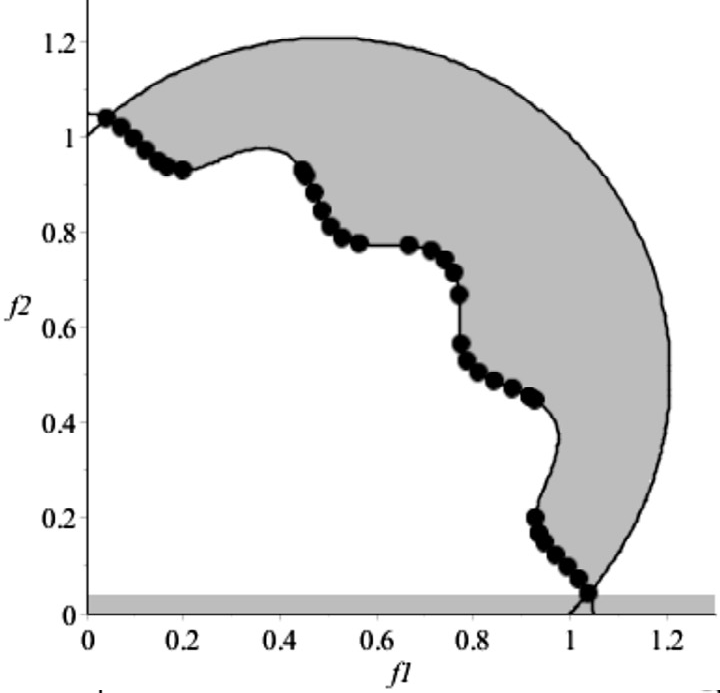}
      \caption{mCHIM}
    \end{subfigure}
    \caption{Strong Pareto Front for Case III. All dominated points are removed from set after applying the Pareto filter.}
    \label{fig:pareto3}
\end{figure}

This problem was proposed in \cite{tanaka1995ga}. Jointly minimize
\begin{align*}
    &f_1(x_1,x_2) = x_1 \\
    &f_2(x_1,x_2) = x_2\\
    &\text{s.t.} \quad g_1(x_1,x_2)= (x_1-0.5)^2 + (x_2-0.5)^2 \leq 0.5\\
    & g_2(x_1,x_2)= x_1^2 + x_2^2 - 1 - 0.1 \cos (16 \arctan (\frac{x_1}{x_2})) \geq 0\\
    &g_3,g_4:0 \leq x_1, x_2 \leq \pi
\end{align*}
This form is convex in $f_1,f_2$ but the non-convex constraints in $g_1,g_2$ forces the Pareto front to be non-convex. While NBI (without Pareto filter) fails in this scenario, both mCHIM and PK extracts the non-dominated Pareto points with limited density $(\sim 40)$. HNPF extracts point with higher density (\textbf{Fig. \ref{fig:pareto3}}). Since the front is strongly affected by the constraints, Fig. \ref{fig:pareto3}(a) shows a sinusoidal {\em weak} Pareto front. %In the absence of the restrictive constraints $g_{1}$ and $g_{2}$, the Pareto solution set is just one point $(0,0)$. 
To arrive at the non-dominated Pareto set, we then post-process this result using the efficient Pareto filter proposed in sub-section \ref{sec:filter}. The updated discontinuous set of non-dominated Pareto points can be seen in Fig. \ref{fig:pareto3}(b) following the visual explanation in Fig. \ref{fig:par_filter}.

\subsection{Case IV: n=3, k=3, m=4}

\begin{figure}[ht]
    \centering
     \begin{subfigure}{0.32\linewidth}
      \centering
      \includegraphics[width=\linewidth]{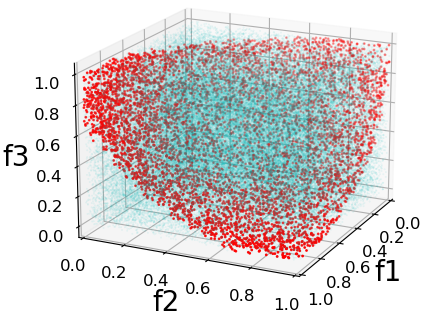}
      \caption{Function Domain}
    \end{subfigure} \qquad
    \begin{subfigure}{0.32\linewidth}
      \centering
      \includegraphics[width=\linewidth]{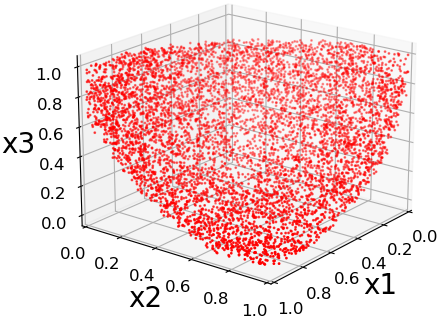}
      \caption{Variable Domain}
    \end{subfigure}
    \caption{Pareto Front for Case IV}
    \label{fig:pareto5}
\end{figure}

This problem was proposed in \cite{ghane2015new}. Jointly minimize
\begin{align*}
    &f_1(x_1,x_2,x_3) = x_1 \\ 
    &f_2(x_1,x_2,x_3) = x_2 \\ 
    &f_3(x_1,x_2,x_3) = x_3 \\
    &\text{s.t.} \quad g_1(x_1,x_2,x_3)= (x_1-1)^2 + (x_2-1)^2 + (x_3-1)^2 \leq 1.0\\
    &g_2,g_3,g_4: x_1,x_2,x_3 \geq 0
\end{align*}
This form is convex in $f_1,f_2,f_3$ but the non-convex constraint in $g_1$ forces the Pareto front to be non-convex. The results using our method, as shown in Fig. \ref{fig:pareto5}, are in good agreement with mCHIM and PK methods with a higher point density.

\subsection{Case V: n=30,k=2,m=30}

This problem was proposed in \cite{zhang2008multiobjective}. Jointly minimize
\begin{align*}
    &f_1(x) = x_1 + \frac{2}{|J_1|}\sum_{j \in J_1}y_j^2 \\
    &f_2(x) = 1 - \sqrt{x_1} + \frac{2}{|J_2|}\sum_{j \in J_2}y_j^2 \\
    &\text{s.t.} \quad g_1,\ldots,g_{30}: 0 \leq x_1 \leq 1, -1 \leq x_j \leq 1, j=2,\ldots,m\\
    & J_1=\{j|j \, \textrm{is odd},2 \leq j \leq m\},J_2=\{j|j \, \textrm{is even},2 \leq j \leq m\}\\
    & y_j = \left\{\begin{matrix}
            x_j - [0.3x_1^2 \cos(24\pi x_1 + \frac{4j\pi}{m}) + 0.6x_1] cos(6\pi x_1 + \frac{j\pi}{m}) \quad j \in J_1   \\ 
            x_j - [0.3x_1^2 \cos(24\pi x_1 + \frac{4j\pi}{m}) + 0.6x_1] cos(6\pi x_1 + \frac{j\pi}{m}) \quad j \in J_2
\end{matrix}\right.
\end{align*}

\begin{figure}[ht]
    \centering
     \begin{subfigure}{0.32\linewidth}
      \centering
      \includegraphics[width=\linewidth]{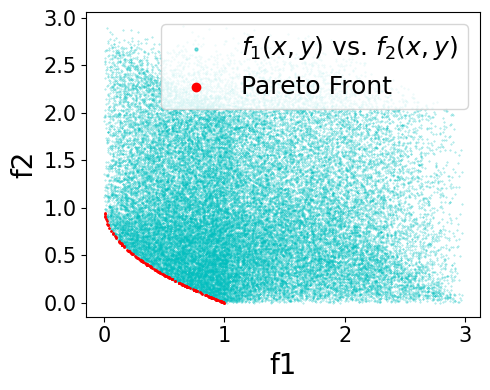}
      \caption{Function Domain}
    \end{subfigure} 
    \begin{subfigure}{0.32\linewidth}
      \centering
      \includegraphics[width=0.975\linewidth]{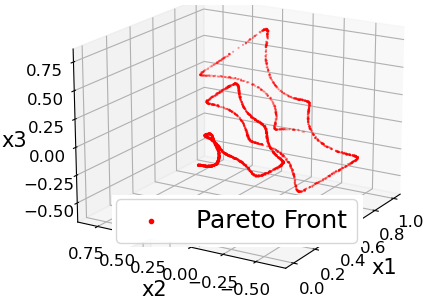}
      \caption{Variable Domain}
    \end{subfigure}
    \begin{subfigure}{0.32\linewidth}
      \centering
      \includegraphics[width=0.7\linewidth]{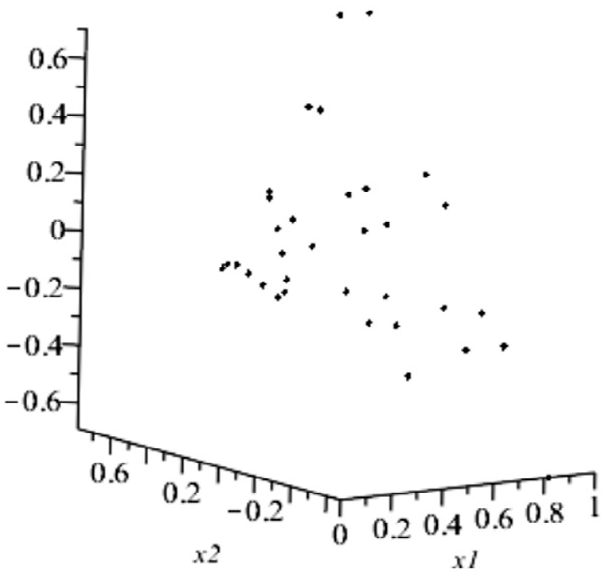}
      \caption{mCHIM}
    \end{subfigure}
    \caption{Pareto Front for Case V. Note the density difference between HNPF and mCHIM in the variable space.}
    \label{fig:pareto7}
\end{figure}

This form is non-convex in both $f_1,f_2$. The dimension of the design variable space is $m=30$. The corresponding Pareto front is non-convex. The results using our method, as shown in Fig. \ref{fig:pareto7}, are in good agreement with mCHIM and PK methods

\subsection{Summary of Results}
%For the numerical experiments presented above, we now summarize these results and present an overall analysis compared to other methods. 

Linear Scalarization, as in \textbf{WSM} \cite{cohon2004multiobjective}, is a well-known approach for Pareto set detection in Fairness and Classification literature. This approach \textbf{fails for all but one (Case I)} of the cases presented above, since either the functions or constraints or both are non-convex. This raises serious concerns regarding validation in Fairness literature: whether the points in the solution set are Pareto optimal or not. If not, then all such works are generating points which are \textbf{non-Pareto} (\textit{weak, strong or otherwise}) in any sense of the definitions posed in Section \ref{sec:definition}. Case III highlights the fallacies of such convexity assumptions, where in spite of functions being convex themselves, the analytical front is non-convex due to the interaction of the functions and non-convex constraints. 

\begin{table}[ht]
    \centering
    % \resizebox{\columnwidth}{!}{%
    \begin{tabular}{c|rrr|rrr}
    \toprule
        & \multicolumn{3}{c|}{\bf HNPF} & \multicolumn{3}{c}{\bf mCHIM} \\ 
        Case & Density & Points & Evals  & Density & Points & Evals  \\ \midrule
        Case II  & 1.83 & 1648 & 90K & 4.39e-2 & 33 & 75,152  \\
        Case III  & 1.37 & 1241 & 90K & 1.01e-2 & 33 & 328,375 \\
        Case IV & 6.57 & 5915 & 90K & 5.86e-3 & 43 & 733,752 \\
        Case V & 0.20 & 184 & 90K & 1.38e-6 & 33 & 2,379,459,895 \\ \bottomrule
    \end{tabular}%}
    \caption{Pareto optimal point density \% (ratio of \#extracted optimal points to \#function evaluations). HNPF finds many more optimal points with many fewer function evaluations. Case I has not been considered by mCHIM, hence left out.}
    \label{tab:eval-mchim}
\end{table}

\textbf{NBI} \cite{das1998normal} works for cases where the detected {\em weak} Pareto front consists of non-dominated points. Therefore, NBI generates correct solution sets in Cases I, II, IV, V with equal density of points on the Pareto front. In essence, applying the Pareto filter on the NBI generated solution set would resolve the discontinuous cases too. 

\textbf{Gobbi} \cite{gobbi2015analytical}, a Genetic Algorithm solution strategy, works for Case I. Their algorithm is developed only for cases where all the functions and constraints are convex. 

NBI, mCHIM,PK and HNPF produce \textbf{only} Pareto points, which is not guaranteed by WSM \cite{cohon2004multiobjective}. Additionally, HNPF generates Pareto points \textbf{uniformly} with \textbf{high} density, while state-of-the-art \textbf{mCHIM} \cite{ghane2015new} and \textbf{PK} \cite{pirouz2016computational}, although accurate, limit themselves to low point density ($\sim 40$) with large computational overhead as the variable dimension scales. \textbf{Table \ref{tab:eval-mchim}} shows a comparison between HNPF and mCHIM. See \textbf{Appendix \ref{app:eval}} for a similar comparison against PK.

\subsection{Runtime Comparison}

\begin{figure}[ht]
    \centering
    \includegraphics[width=0.5\linewidth]{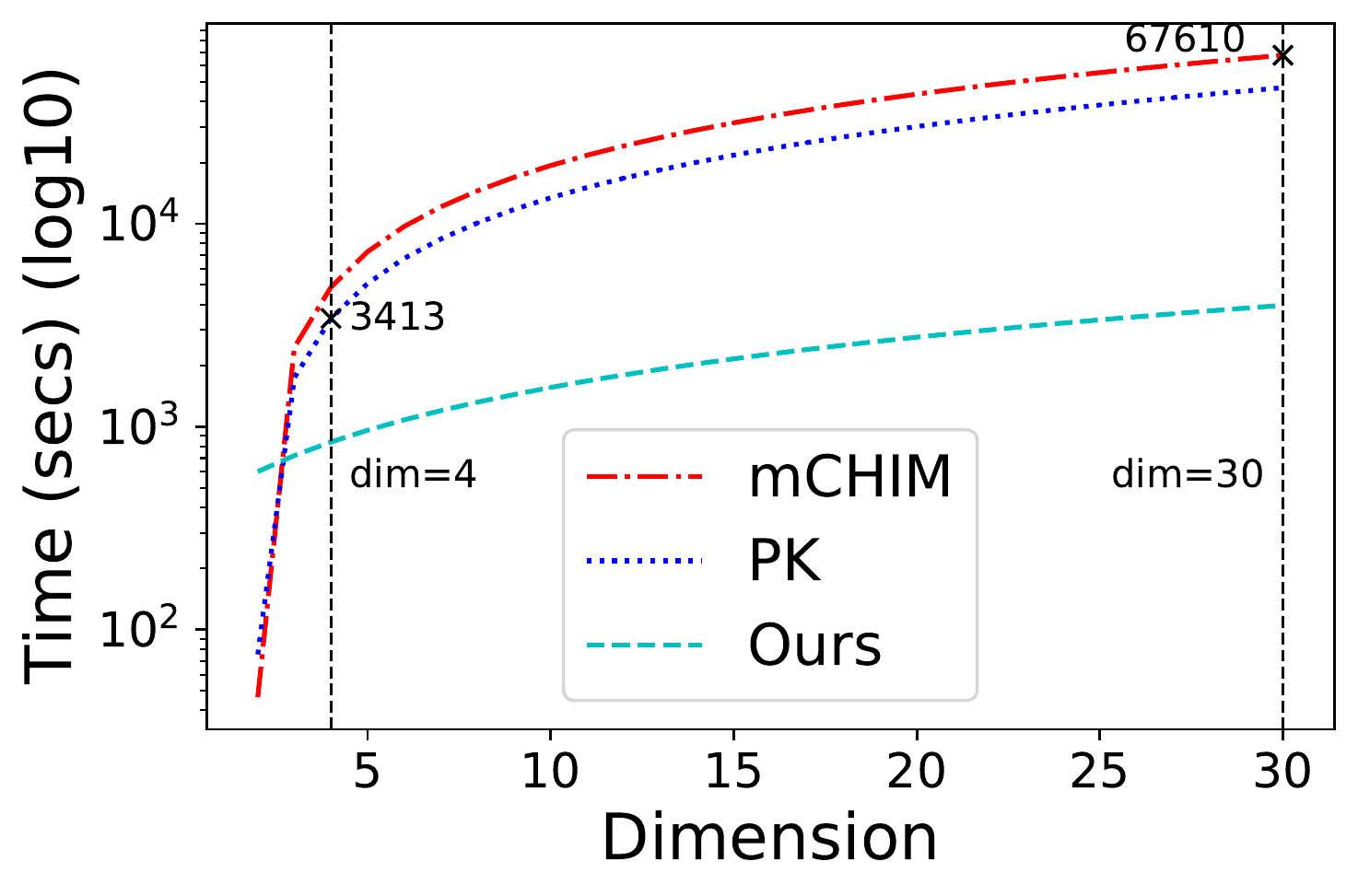}
    \caption{Runtime of HNPF \vs mCHIM and PK, as the variable dimension increases. All methods have a linear increase in runtime with dimension, but HNPF scales much better.}
    \label{fig:runtime}
\end{figure}

Using numerical experiments, we previously verified that mCHIM, PK and our method always arrives at the correct results for all the considered cases. We now perform a compute time analysis against mCHIM and PK, to demonstrate improved performance using our proposed approach. The trajectories in \textbf{Fig. \ref{fig:runtime}} show the compute times for the high dimensional Case VII. Note that for mCHIM and PK, the timings are reported for dimensions n=30 and n=4, respectively. For our method, the runtimes are reported for Case VII with  the variable space dimension ranging from $[2-30]$.

The reported runtime with two dimensional variable space might give the false notion that mCHIM and PK are more efficient than HNPF. However, as the variable dimension increases, both mCHIM and PK become far more expensive, as shown in Fig. \ref{fig:runtime}. %, HNPF seems to be more expensive for lower dimensional cases (upto 2) however, beyond 2 PK and mCHIM methods incur substantial compute overheads. 
These methods also produce a low density of Pareto points $(\sim 40)$, while HNPF yields high density $(\sim 1k)$. %We would like to emphasize that while accuracy is of prime importance, in addition to it run-time scalability with increasing problem dimension is equally important for practical applications. 
Since both mCHIM and PK are based on enhanced scalarization, solving the resulting problem to extract Pareto points suffers from scaling issues.

\subsection{Loss Profile} \label{sec:profile}

\begin{figure}[ht]
    \centering
    \begin{subfigure}{0.24\linewidth}
      \centering
      \includegraphics[width=\linewidth]{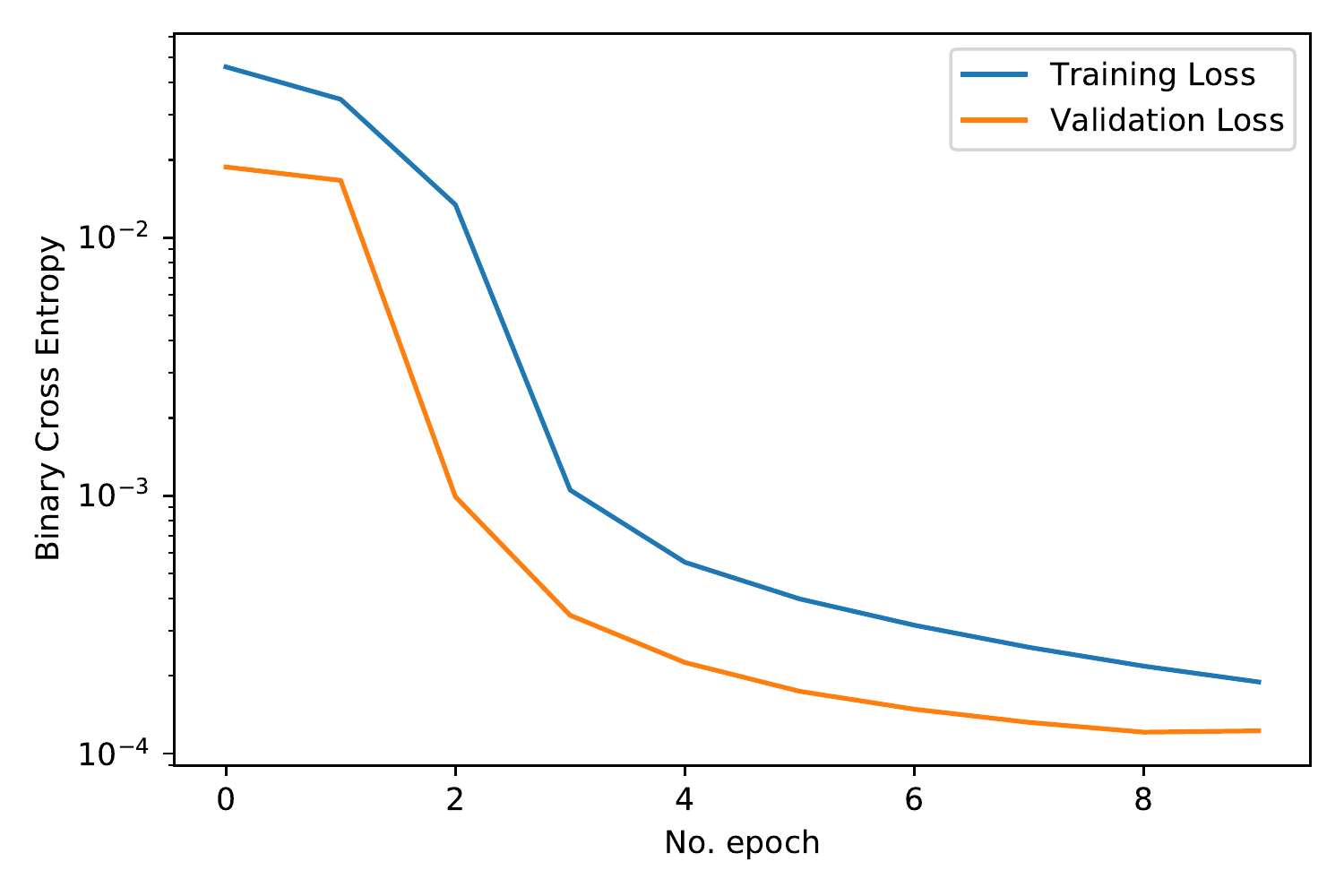}
      \caption{Case II}
    \end{subfigure}
    \begin{subfigure}{0.24\linewidth}
      \centering
      \includegraphics[width=\linewidth]{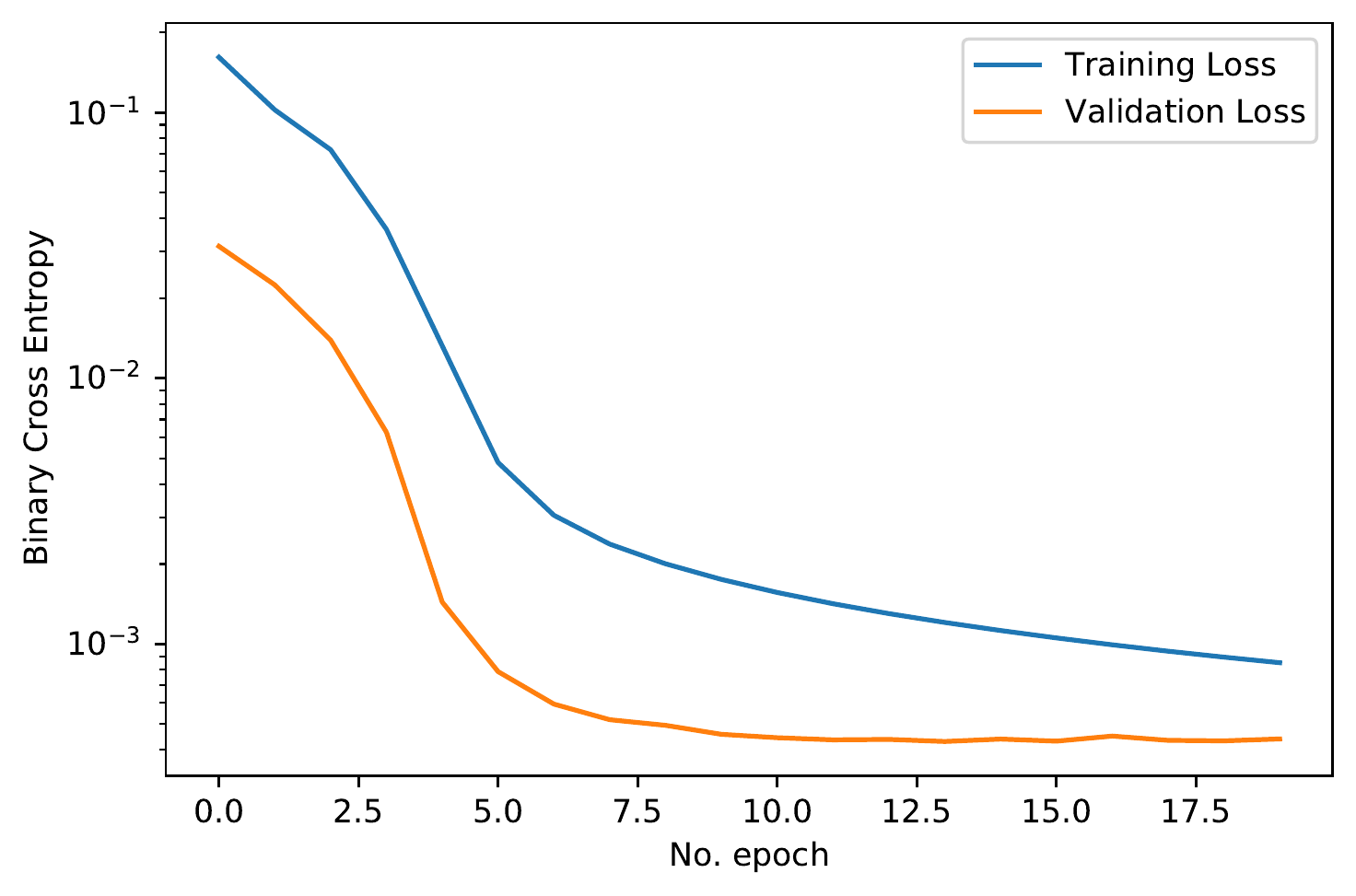}
      \caption{Case III}
    \end{subfigure}
    \begin{subfigure}{0.24\linewidth}
      \centering
      \includegraphics[width=\linewidth]{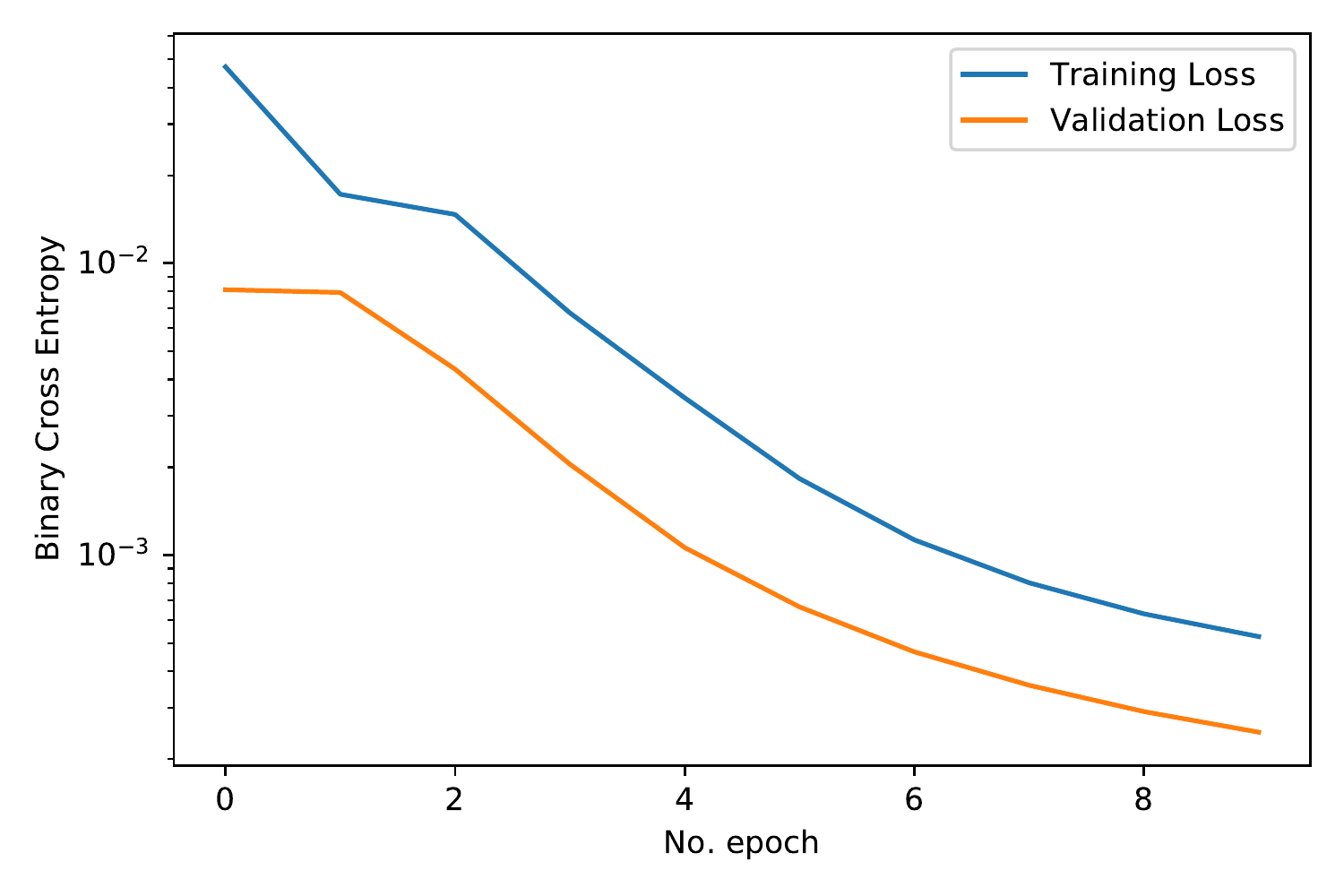}
      \caption{Case IV}
    \end{subfigure}
    \begin{subfigure}{0.24\linewidth}
      \centering
      \includegraphics[width=\linewidth]{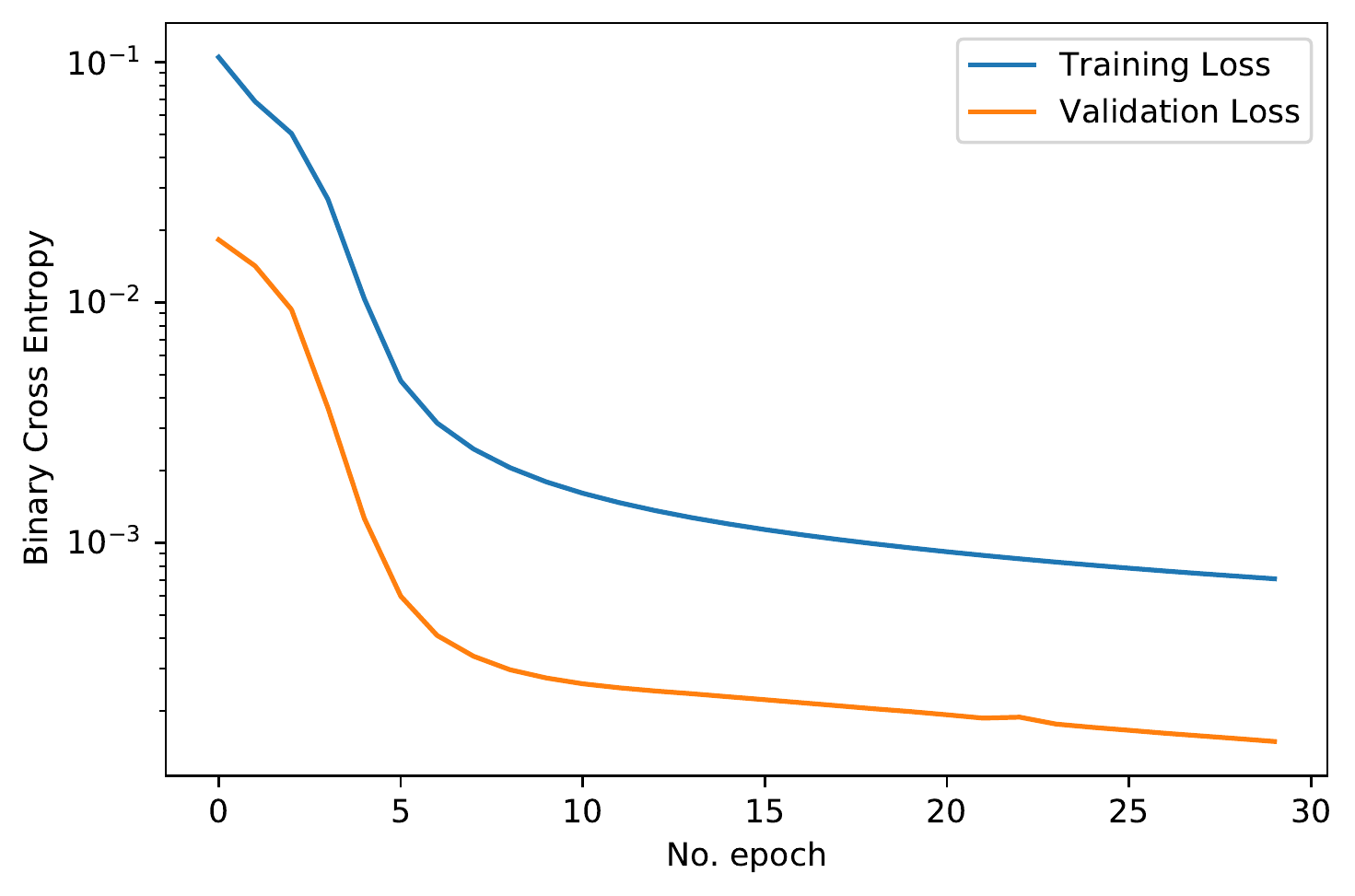}
      \caption{Case V}
    \end{subfigure}
    \caption{Training (blue) and validation (orange) curves for 4 cases. An increase in the dimensions of the design variable space results in increased costs for constructing the L matrix. Consequently, the network takes more epochs to converge.}
    \label{fig:loss}
\end{figure}

We now briefly discuss the training process for the cases shown above. On an average, the network takes around $10/20/30$ epochs for the simple/moderate/hard cases as visualized in \textbf{Fig. \ref{fig:loss}}. Since the last layer of the network is classifying points as being {\em weak} Pareto or not, the runtime is dictated by the complexity of the curve in the design variable space. The more non-linear the solution manifold, the more training time is required to approximate it. 

Case II and IV both converge within 10 epochs although they lie in 2D and 3D space, respectively. Per Fig. \ref{fig:pareto2} and \ref{fig:pareto5} (b), the design variable space is convex and so the solution manifold is less complicated. Although in 2D variable space, Case III takes 20 epochs owing to the sinusoidal solution manifold. Case V converges in 30 epochs, the design space is 30 dimensional, hence the compute complexity increases due to the construction of a larger $L$ matrix. The validation loss curve lies below the training loss (but strictly at scale), suggesting that our low-weight network did not over/underfit. 

\section{Modeler Interpretability}

\begin{figure}[ht]
    \centering
    \includegraphics[width=0.4\linewidth]{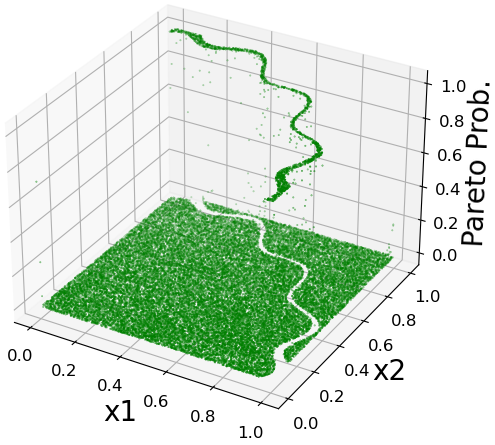}
    \caption{Classification boundary of {\em weak} Pareto points for Case III. The final layer assigns a probability score to each point in the variable space as being Pareto optimal or not. A modeler needs to examine the produced points/manifold against the known analytical form in the variable domain to verify the correctness of the network's working.}
    \label{fig:class}
\end{figure}

Motivated by \citet{lipton2018mythos}'s definitions of model interpretability and trust, we adopt the persona of a modeler in assessing the interpretability of our model. In all of the problems above, the approximate manifold $\tilde{M}$ is described by the user specified loss function. If a domain specific analytical solution ($M(X^*)=0$) is known, then the approximate network generated solution set ($\tilde{M}(\tilde{X})=0$) can be verified by comparing $\tilde{X}$ and $X^*$. Additionally, a domain-specific modeler can also compare the approximate manifold $\tilde{M}$ (see \textbf{Fig. \ref{fig:class}} for case III), at the last layer of the network, against the true manifold $M$ known from the analytical form. If the modeler is able to verify that the network classifies the correct (truly Pareto optimal) data points in the variable space as being Pareto optimal (high probability value), the trust in the network's working is established.  

%%%%%%%%%%%%%%%%%%%%%%%%%%%%%%%%%%%%%%%%%%%%%
%%%%%%%%%%%%%%%%%%%%%%%%%%%%%%%%%%%%%%%%%%%%%
%%%%%%%%%%%%%%%%%%%%%%%%%%%%%%%%%%%%%%%%%%%%%

\section{An Application: Fair Search}

Imagine a new policy for predictive policing is under consideration, with various public arguments being published for and against adoption. By reviewing this body of arguments, one might arrive at an informed and balanced understanding of the issue and public debate surrounding it. This, in turn, could guide citizens or lawmakers in voting, or help a journalist to provide balanced reporting. %news coverage of the evolving debate. 

Assume a search engine is used to find information that is both relevant and balanced.  Specifically, assume we wish to maximize two objective functions computed on the set of retrieved search results: 1) relevance and 2) diversity (i.e., balanced inclusion of search results that are for and against the proposed policy). Note that our diversity target is specified here as a soft objective function to maximize rather than a hard constraint to rigidly enforce. Let us make a further independence assumption between relevance and diversity: knowing that a retrieved document is relevant does not provide any indication as to whether or not that document presents an argument for or against the proposed policy.

\citet{gao2019fair} present such a fair search problem as follows (albeit with a different motivating back story). Let $S$ denote the search result set of cardinality $|S| = s$. Assume each document $d_i$ has binary relevance $r_i\in\{0,1\}$ and group assignment $g_i\in\{0,1\}$ (i.e., for or against the policy, in our scenario), and that $r_i$ and $g_i$ are independent, as above. The optimization goal is dual maximization of the average relevance $f_r = \overline{r_i} = \frac{\sum r_i}{n}$ and the entropy $f_g = H(\frac{\sum g_i}{n})$ of the search result set, with entropy used as the measure of diversity ({\em aka} parity, balance, or group fairness). 

\subsection{Insights from Pareto Framing} 

\citet{gao2019fair} pose several questions, including ``1. What are the possible relevance and fairness scores of a solution set $S$ (the solution space)?'' and ``2. What is the trade-of between fairness and relevance?'' They define the solution space as the set of all possible subsets $S \subset D$ in document collection $D$ having cardinality $|S| = s$. They then proceed to investigate these questions via simulation: generating different subsets of search results and inspecting the empirical distributions of scores. In contrast, we suggest a conceptual Pareto framing provides more direct and informed answers. 

While we have emphasized generality of Pareto optimality under competing objectives with constraints, Pareto optimality is of course also applicable to simpler optimization problems, such as posed here. Firstly, since there are no constraints on the solution space, the feasible set spans the full range of $f_r\in[0,1]$ and $f_g\in[0,1]$. Secondly, since relevance and entropy are independent, they do not compete: maximizing one does not preclude maximizing the other, and each can be considered separately in turn. Relevance is maximized when all search results are relevant ($f_r=1$), while entropy is maximized when search results are evenly split across the two groups ($f_g=1$). This yields {\em weak} Pareto fronts at $f_r=1$ and $f_g=1$, with the only strong Pareto solution at the intersection of both fronts ($f_r=f_g=1$), when search results are all relevant and are evenly split across the two groups being represented. 

With regard to the probability of observing any given ($f_r,f_g)$ score for a given search result set $S$ (i.e., the chance of achieving the optimum $f_r=f_g=1$ or any other feasible point), since relevance and entropy are independent, their joint distribution is defined simply by the product of probability distribution functions (PDFs) for relevance $P(r_i)=p_r$ and group membership $P(g_i)=p_g$. In practice, we must induce these PDFs from data, but this is standard estimation and not unique to this particular problem setting. Moreover, this permits analytical analysis without simulation.

Finally, whereas \citet{gao2019fair}'s optimization problem is relatively easy (no constraints or competing objectives), one could easily introduce further complications.  For example, imagine public opinion is highly skewed, such that nearly all relevant information supports one side of the argument.  In this scenario, in order to get more balanced coverage of the minority position, we would need to include more non-relevant search results, forcing the user to sift through a larger result set to find relevant and balanced information. As a second example, given controversy surrounding predictive policing policies, one might like to constrain search results to enforce racial parity across authors of retrieved documents in the results set.  In either case, our Pareto optimality framing would allow principled reasoning about the resulting solution spaces.

\subsection{Another Test Case for HNPF} \label{sec:test} 
As in Section \ref{sec:results}, evaluating our HNPF approach on solutions with known analytical forms allows us to verify its accuracy.  As discussed above, we know from first principles that the fair search problem considered here has {\em weak} Pareto fronts at $f_r=1$ and $f_g=1$, with the only strong Pareto solution at the intersection of both fronts ($f_r=f_g=1$).  By running HNPF on this problem, we can verify it induces these expected fronts as another check on its correctness.

\citet{gao2019fair} simulate $r_i$ and $g_i$ values based on observed statistics of the YOW RSS Feed dataset \cite{zhang2005bayesian}. They consider $n=48$ data points drawn from this dataset, with empirical $\hat{p_r}=0.56$ and and $\hat{p_g}=0.5$. Sampling from  binomial distributions $R \sim B_r(n=48,p_r=0.56)$ and $G \sim B_g(n=48,p_r=0.50)$, in expectation we anticipate $R = (\sum^{n=48} r_i) \approx 27$ relevant documents and an even split of $G = (\sum^{n=48} g_i) = 24$ across groups. Note that whereas average relevance $f_r$ is maximized when all documents are relevant, the entropy $f_g$ is maximized when documents are evenly split by group. The probability distributions above will thus naturally tend to yield near optimal entropy $f_g \approx 1$ (with $p_g=0.5$) but only mediocre average relevance $f_g \approx \overline{r_i}=0.56$. With $p_r = 0.56$, the probability of maximum relevance $P(f_r=1) = P(R=48) \approx 10^{-60}$. %, so exceedingly low.

\begin{figure}[ht]
    \centering
     \begin{subfigure}{0.4\linewidth}
      \centering
      \includegraphics[width=\linewidth]{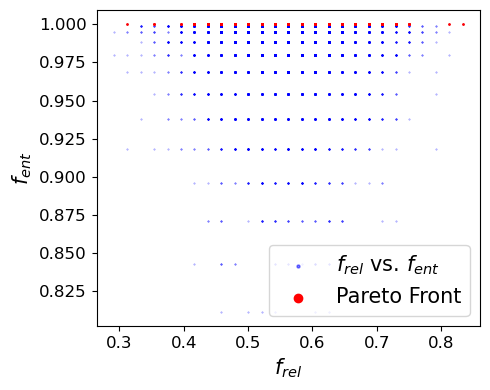}
      \caption{Function Domain}
    \end{subfigure}\qquad
    \begin{subfigure}{0.4\linewidth}
      \centering
      \includegraphics[width=\linewidth]{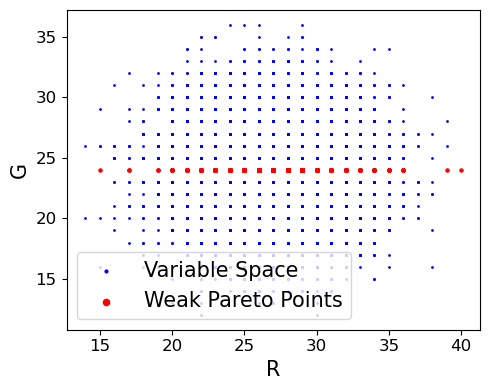}
      \caption{Variable Domain}
    \end{subfigure}
    \caption{Weak Pareto Front for RSS dataset}
    \label{fig:paretorss}
\end{figure}

{\bf Figure \ref{fig:paretorss}} shows the {\em weak} Pareto front for (a) the function domain $(y=f_r,x=f_g)$ and (b) the variable domain $(x=R, y=G)$. As expected, we see the sample distribution lays roughly symmetrically about the expectation in both $R$ and $G$ in the variable domain, yielding near optimal entropy $f_g$ and mediocre average relevance $f_r$ in the function domain. Given this sample, the network correctly identifies the Pareto front for entropy at $f_g=1$, corresponding to $G=24$ in the variable domain.  Note that this is still a {\em weak} front, where all points to the left are dominated by those to the right, hence the need for the Pareto filter to identify the non-dominated set. However, the network cannot identify the Pareto front for relevance at $f_r=1$ due to sample sparsity relative to the sampling distribution. As noted above, with $p_r=0.56$, $P(f_r=1) \approx 10^{-60}$, suggesting we would need $10^{60}$ samples to observe $f_r=1$. 

Our results above follow \citet{gao2019fair} in sampling from the variable domain according to ($p_r,p_g$). This makes sense if we want to explore the solution space via simulation, as they do. However, if our goal is actually to identify Pareto fronts (e.g., to measure how far a given solution is from optimality), we can instead probe the solution space far more efficiently by uniformly sampling the variable domain ($r,g\in[0,1]$). \textbf{Appendix \ref{app:rss}} presents these results. %for the extracted front under a uniform distribution.

\section{Conclusion}

A hybrid, two-stage, neural-Pareto filter based optimization framework is presented for extracting the Pareto optimal solution set for multi-objective, constrained optimization problems. The proposed method is computationally efficient and scales well with increasing dimensionality of design variable space, objective functions, and constraints. Results on verifiable benchmark problems show that our Pareto solution set accuracy compares well with existing state of the art NBI, mCHIM, and PK methods. The proposed neural architecture is fully interpretable with a Fritz-John conditions inspired discriminator for {\em weak} Pareto manifold classification. We also show that the approximation error between the true and extracted Pareto manifold can be easily verified for analytical solutions. %As a future extension, we would like to extend this work to fairness and diversity guided classification tasks for accuracy trade-offs. In our future work, we will extend this approach with an adaptive point generation strategy to further improve the compute efficiency.

%\bibliographystyle{plainnat}
%\bibliography{References}

\clearpage

\appendix

\section{Additional Cases} \label{app:cases}

\subsection{Case VI: n=2, k=2, m=5}

This problem was proposed in \cite{dutta2011new}. Jointly minimize
\begin{align*}
    &f_1(x_1,x_2) = x_1 \\
    &f_2(x_1,x_2) = x_2\\
    &\text{s.t.} \quad g_1(x_1,x_2)= (x_1-0.5)^2 + (x_2-0.5)^2 \leq 0.5\\
    & g_2(x_1,x_2)= x_1^2 + x_2^2 - 1 - 0.1 \cos (16 \arctan (\frac{x_1}{x_2})) \geq 0\\
    & g_3(x_1,x_2) = max(|x_1-0.6|,|x_2-0.7|) - 0.2 \geq 0\\
    &g_4,g_5:0 \leq x_1, x_2 \leq \pi
\end{align*}

\begin{figure}[ht]
    \centering
     \begin{subfigure}{0.32\linewidth}
      \centering
      \includegraphics[width=\linewidth]{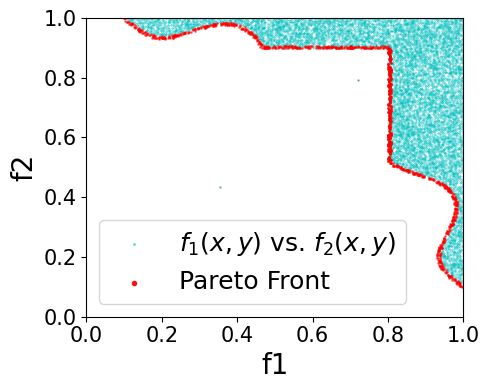}
      \caption{Dominated}
    \end{subfigure} \qquad
    \begin{subfigure}{0.32\linewidth}
      \centering
      \includegraphics[width=\linewidth]{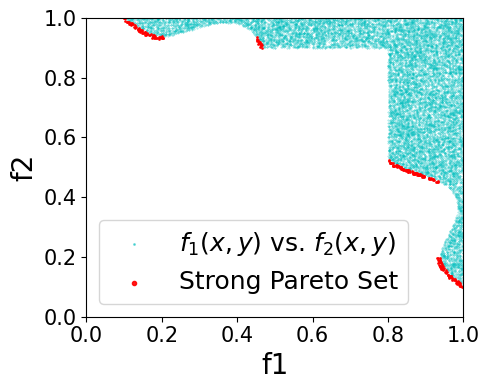}
      \caption{Non-Dominated}
    \end{subfigure}
    \begin{subfigure}{0.32\linewidth}
      \centering
      \includegraphics[width=\linewidth]{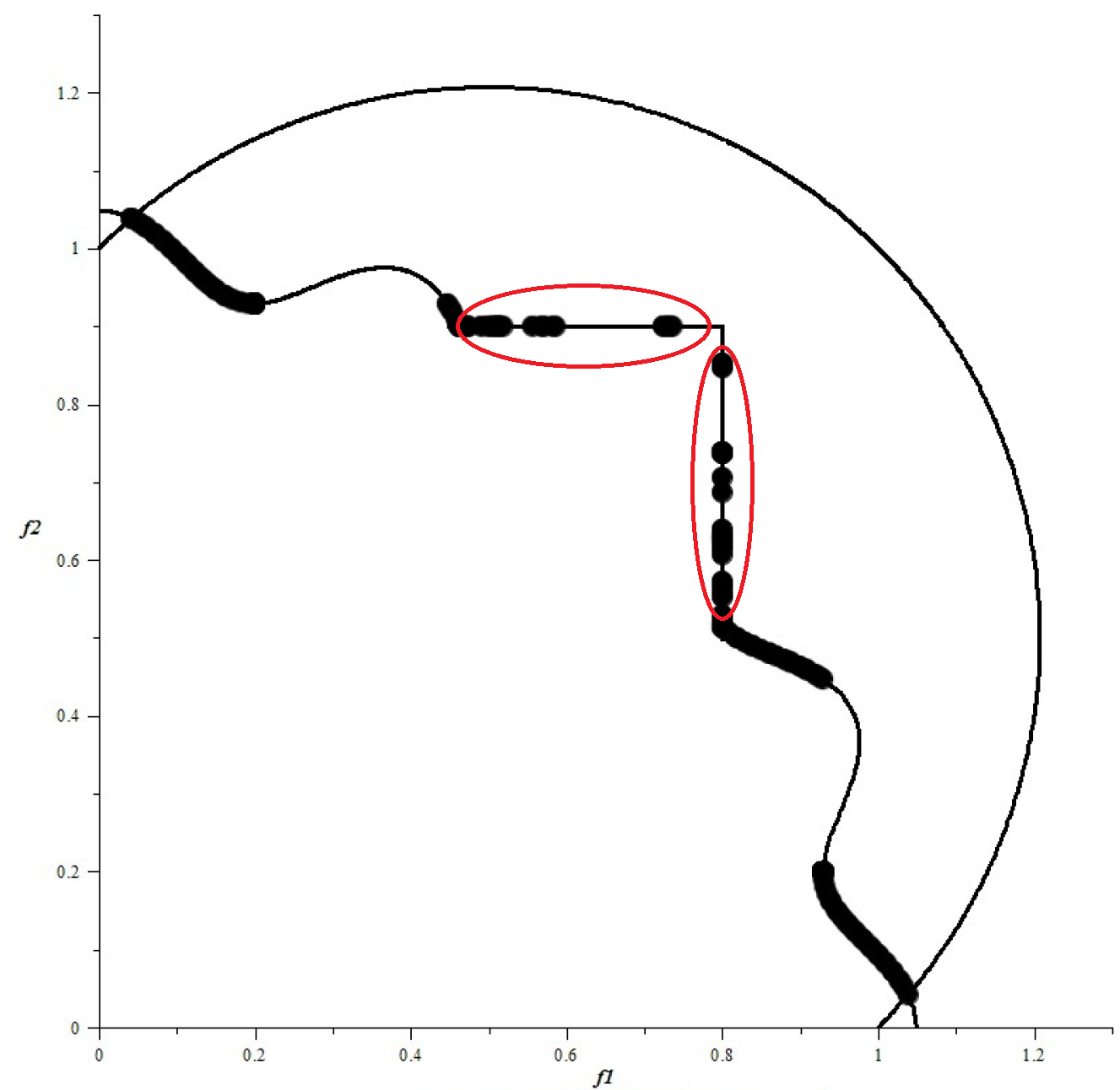}
      \caption{PK}
    \end{subfigure} \qquad
    \begin{subfigure}{0.32\linewidth}
      \centering
      \includegraphics[width=\linewidth]{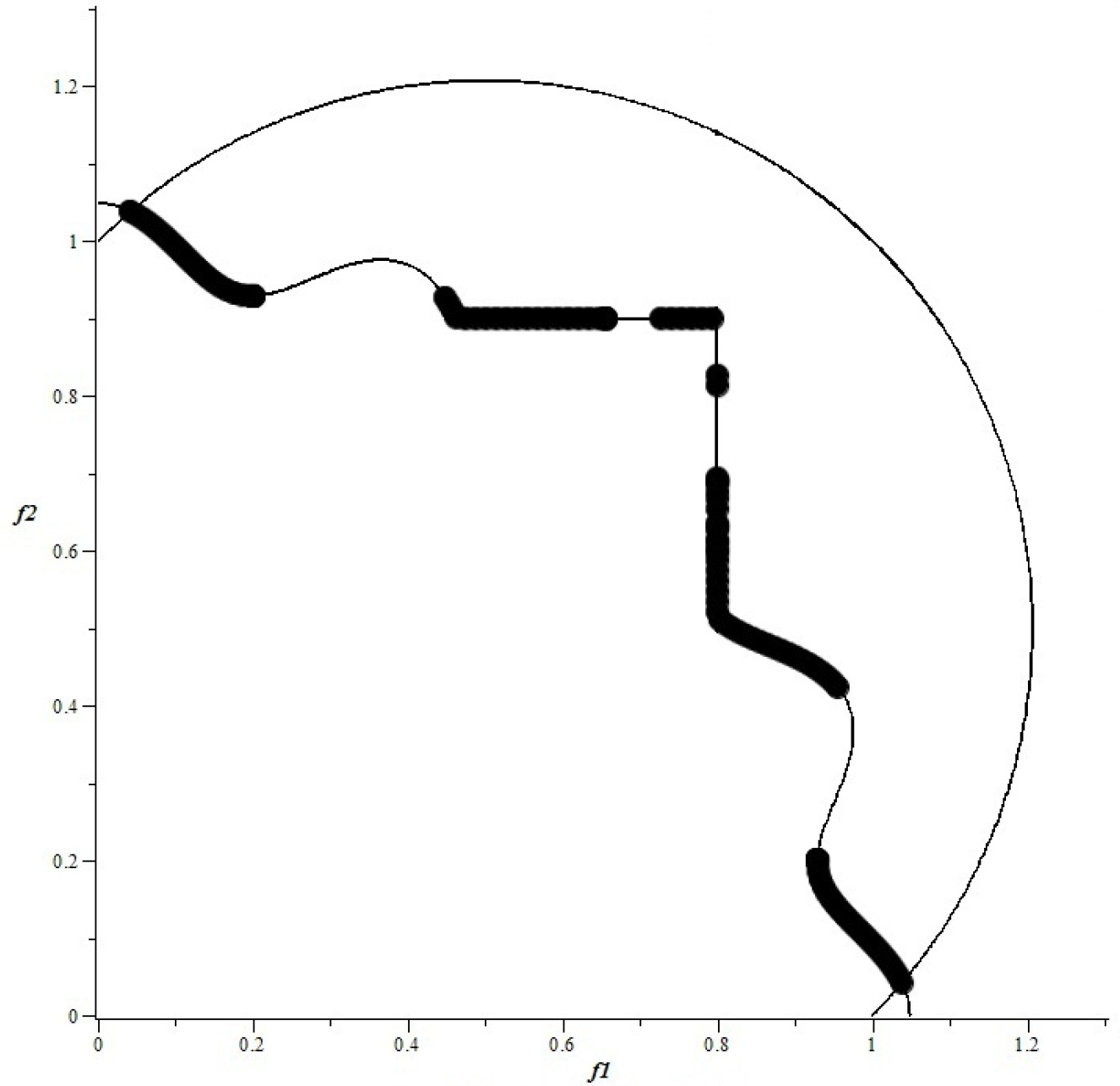}
      \caption{NBI}
    \end{subfigure}
    \caption{Strong Pareto Front for Case VI. Note that all the dominated points are removed from the set after application of Pareto filter. Also note the low density of points in PK and the even spread of points in NBI. Also note that PK is not able to remove some of the dominated points on the horizontal and vertical lines.}
    \label{fig:pareto4}
\end{figure}

This form is an extension of Case III, with an additional \textit{max} boundary constraint $g_3$. Both mCHIM and PK computes the true Pareto front with limited density $n=40$ points. Fig. \ref{fig:pareto4} (a) shows the {\em weak} Pareto front with dominated points. As before, after post-processing with the proposed Pareto filter we arrive at the Pareto set with non-dominated points shown in Fig. \ref{fig:pareto4} (b).

\subsection{Case VII: n=30, k=2, m=30}

This problem was proposed in \cite{pirouz2016computational}, albeit with a  discrepancy\footnote{Although the normalization term proposed in $f(x)$ is $m-1$, it does not generate the curve reported in \cite{pirouz2016computational}. %The shown form is an updated version of the one proposed in \cite{zitzler2000comparison}. 
We were able to replicate the shown curve, in our experiments, by choosing a normalizing constant of $10000$.}. Jointly minimize
\begin{align*}
    &f_1(x) = x_1 \\
    &f_2(x) = f(x)\left (1-\left (\frac{f_1(x)}{f(x)}\right )^{0.5}-\frac{f_1(x)}{f(x)}sin(10\pi x_1) \right )\\
    &\text{s.t.} \quad f(x) = 1 + \frac{9}{m-1} \sum_{i=2}^nx_i^2\\
    & g_1,\ldots,g_{30}:x_1 \in [0,1], x_i \in [-1,1],\forall i =2,\ldots,m
\end{align*}

\begin{figure}[ht]
    \centering
     \begin{subfigure}{0.32\linewidth}
      \centering
      \includegraphics[width=\linewidth]{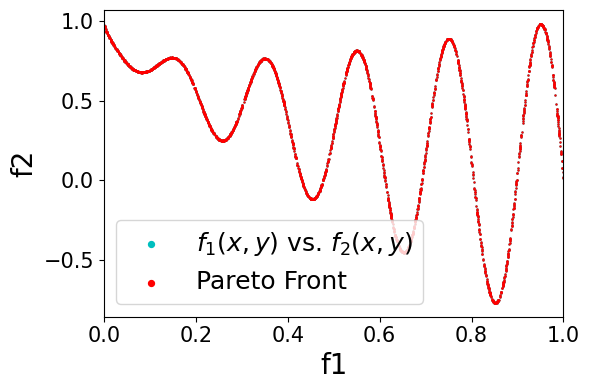}
      \caption{Dominated}
    \end{subfigure} \qquad
    \begin{subfigure}{0.32\linewidth}
      \centering
      \includegraphics[width=\linewidth]{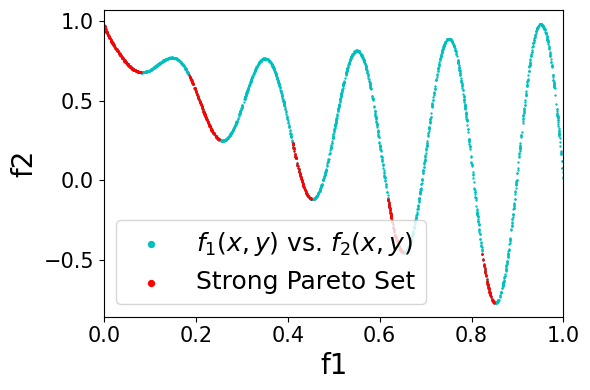}
      \caption{Non Dominated}
    \end{subfigure}
     \begin{subfigure}{0.32\linewidth}
      \centering
      \includegraphics[width=\linewidth]{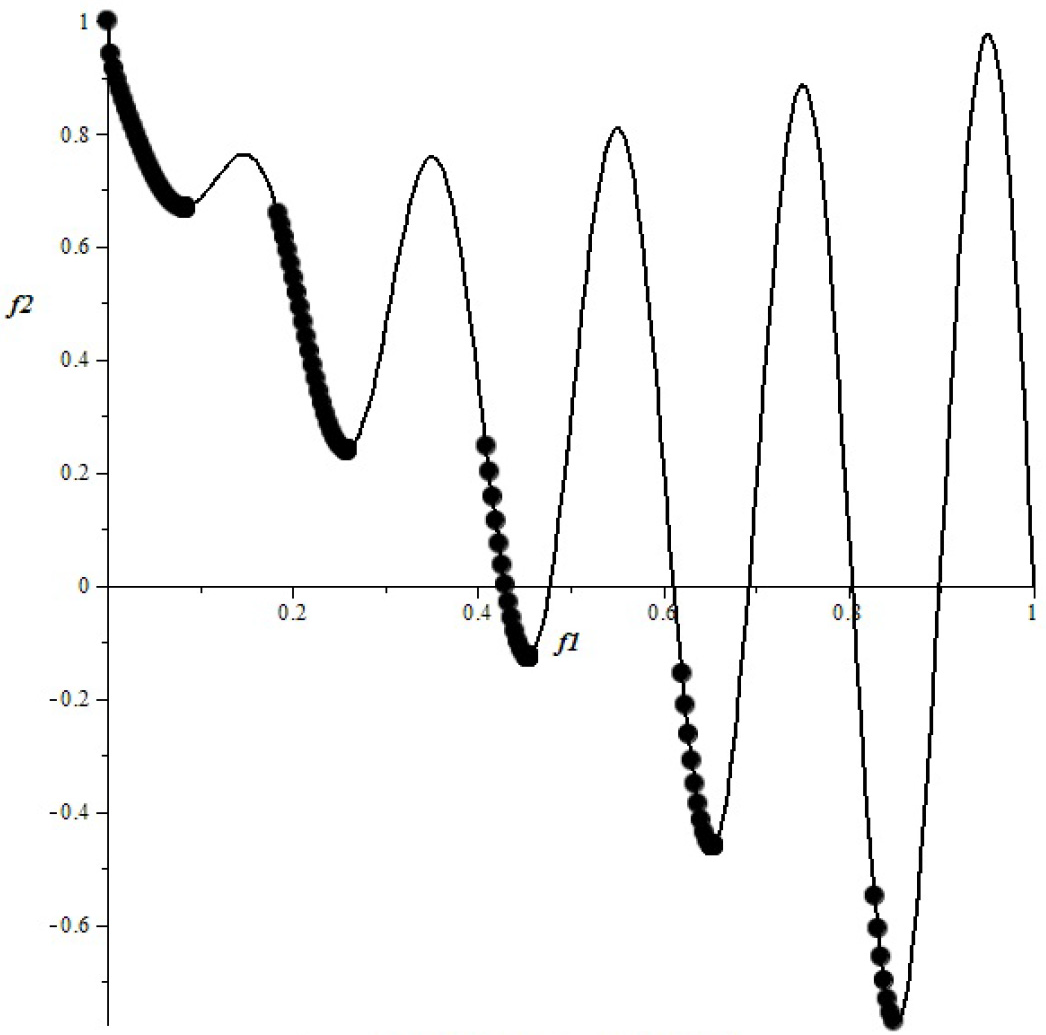}
      \caption{PK}
    \end{subfigure} \qquad
    \begin{subfigure}{0.32\linewidth}
      \centering
      \includegraphics[width=\linewidth]{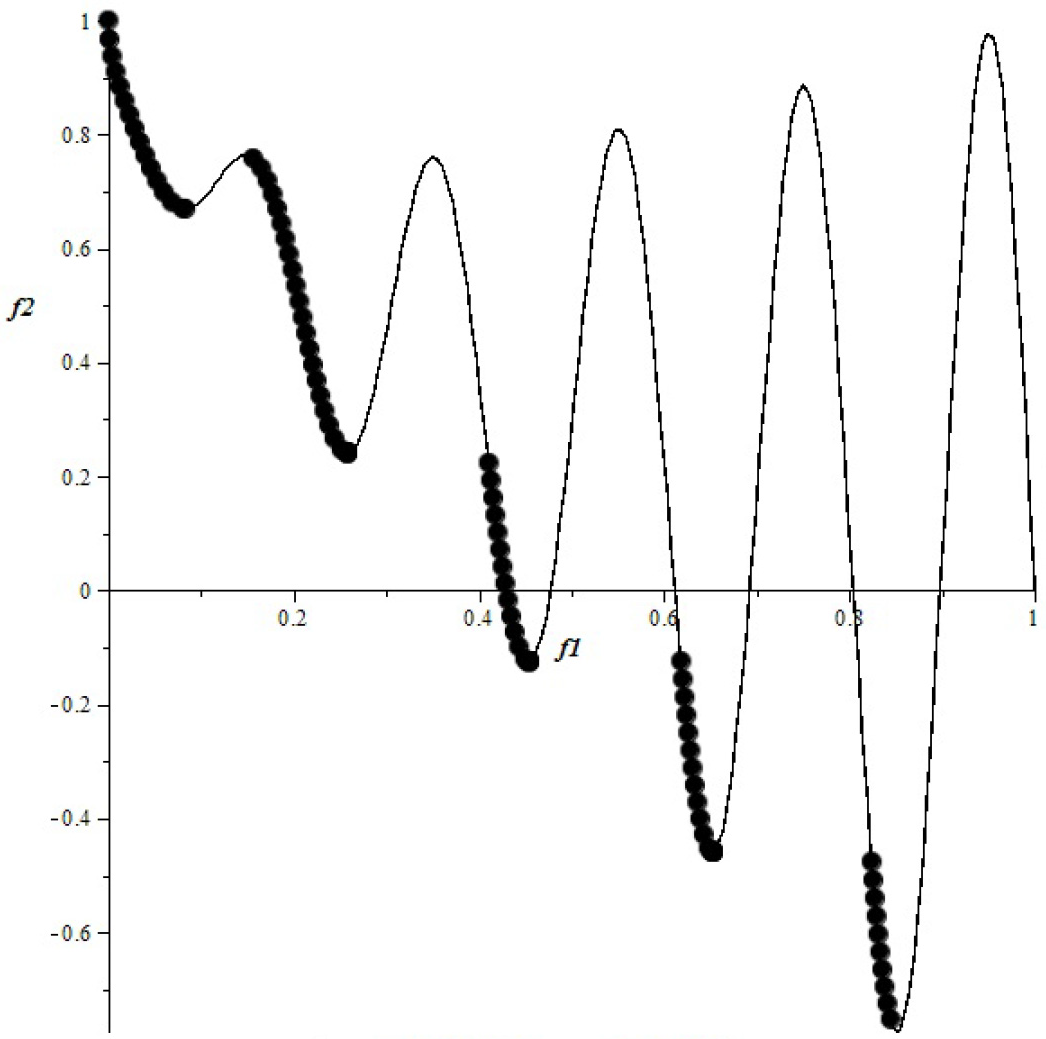}
      \caption{NBI}
    \end{subfigure}
    \caption{Pareto Front for Case VII. Note that all the dominated points are removed from the set after application of Pareto filter. Also note the low density of points in PK and the even spread of points in NBI.}
    \label{fig:pareto6}
\end{figure}

This form is convex in $f_1$ and non-convex in $f_2$. The dimension of the design variable space is $m=30$. The corresponding Pareto front is non-convex. The results using our method, as shown in Fig. \ref{fig:pareto6}, are in good agreement with mCHIM and PK methods. Even in this high-dimensional setting, we obtain a {\em weak} Pareto front with high point density as shown in \textbf{Fig. \ref{fig:pareto6}} (a). The distribution of the objective space in this setting is such that the entire space is the front itself. Hence, we cannot see the cyan points in Fig. \ref{fig:pareto6} (a). As before, after post-processing with the proposed Pareto filter we arrive at the Pareto set with non-dominated points shown in Fig. \ref{fig:pareto6} (b), where the dominated points are now visible.

\section{Working of Pareto filter} \label{app:filter}

The algorithm starts with the set of all {\em weak} Pareto points $p \in P$, which will be refined through the iterative process. The loop (\textbf{line 4}) iterates over all the functions $f_i$. It checks for set of dominating and non-dominating points for all discretization levels (\textbf{line 6}) and appends them to a temporary list (\textbf{line 10}). If multiple points do exist (\textbf{line 11}) for a given level (cardinality $> 1$), then there certainly are dominated points. The non-dominated point is one which has the lowest function value for the next function $f_q, q=i+1$. This (\textbf{line 12}) states that a point which seems non-dominated for a given function $f_i$ might be dominated for other functions $f_r,r \in k, r \neq i$, but will be taken care of when iterating through function $f_r$. Once the non-dominated point has been found, it implies that all the other points are in fact dominated, hence should not be considered for further evaluation. They are rejected (\textbf{line 13}) from the active set $P = P \backslash (temp \backslash x_p)$. The output is the set of strong Pareto points which were all non-dominated for every function $f_i$, and are essentially the points in the {\em weak} Pareto set $P$ that survived the filtering process (\textbf{line 13}) for every $f_i$.

\section{Error Bound} \label{app:error}

For a user-prescribed relaxation margin $0 \leq \epsilon \leq 1$, the approximation error between the network extracted manifold $\tilde{M}(\tilde{X})$ and the true solution $M(X^*)$ is bounded below by  $\|\tilde{M}(\tilde{X}) - M(X^*)\|_2 \leq \epsilon$. Assuming the $L$ matrix from Eq. \eqref{eq:fjmatrix} is square, we have $det(L) = 0$. This form holds for some of the problems chosen in our numerical experiments (Cases I, II, III), where the number of functions and constraints are equal. From Leibnitz formula for determinants:
\begin{align*}
    det(L) = det\Big( \begin{bmatrix}
\nabla F & \nabla G \\
\mathbf{0} & G
\end{bmatrix} \Big)
%&= det \Big( \begin{bmatrix}
%\nabla F & \mathbf{0} \\
%\mathbf{0} & I
%\end{bmatrix} \begin{bmatrix}
% I &  \mathbf{0} \\
%\mathbf{0} & G
%\end{bmatrix} \begin{bmatrix}
% I & \nabla G \\
%\mathbf{0} & I
%\end{bmatrix}  \Big) \\
= det(\nabla F) det(G) = 0
\end{align*}
Further assume that, 
\begin{align}
|det(L(\tilde{x}))| \leq \epsilon, \, \tilde{x} \neq x^* \label{eq:approx}
\end{align}
where $\epsilon > 0$, and $x^*$ and $\tilde{x}$ are the optimal points and the network generated approximate solution points, respectively. The Fritz John necessary conditions in Eq. \eqref{eq:fjcond} for {\em weak} Pareto optimality is:
\begin{align}
    det(L(x^*)) = 0 \label{eq:true}
\end{align}
Combining the assumption in Eq. \eqref{eq:approx} and Eq. \eqref{eq:true}, we have
\begin{align}
|det(L(\tilde{x}))-det(L(x^*))| \leq \epsilon
\end{align}
The solution manifold $M(X)=det(L(X))=0$ is weakly Pareto optimal. We assume a low precision manifold $M(\tilde{x})$ such that:
\begin{align}
    \|M(\tilde{X}) - M(X^*)\|_2 \leq \epsilon \label{eq:netopt}
\end{align}
When the network converges, Eq. \eqref{eq:netopt} will hold for the network approximated $\tilde{M}(x)$. Here, $\tilde{x} \in \tilde{X} =\{x|\tilde{M}(x)=0\}$ and $X^*$ is the set of true optimal points such that $M(x^*) = 0, \forall x \in X^*$. Since we explicitly specify $\epsilon$ in our loss description, we know that the network generated solution is $\epsilon$ close to $M(\tilde{x})$ if:
\begin{align}
    \|M(\tilde{X}) - \tilde{M}(\tilde{X}) \|_{2} \leq C \epsilon, \quad 0 \leq C \leq 1 \label{eq:close}
\end{align}
The form in Eq. \eqref{eq:close} implies that if we are able to find such a $C$, then we implicitly satisfy Eq. \eqref{eq:netopt}. Hence, 
\begin{align}
    \|\tilde{M}(\tilde{X}) - M(X^*)\|_2 \leq \epsilon
\end{align}

\section{Density Comparison} \label{app:eval}

\begin{table}[ht]
    \centering
    % \resizebox{\columnwidth}{!}{%
    \begin{tabular}{c|rrr|rrr}
    \toprule
        & \multicolumn{3}{c|}{\bf HNPF} & \multicolumn{3}{c}{\bf PK} \\ 
        Case & Density & Points & Evals  & Density & Points & Evals  \\ \midrule
        Case V & 1.34 & 1206 & 90K & 2.21e-4 & 100 & 45,126,324 \\
        Case VI & 0.20 & 180 & 90K & 1.32e-2 & 151 & 1,139,781 \\ 
        Case VII & 6.57 & 5915 & 90K & 9.29e-5 & 101 & 108,685,605 \\ \bottomrule
    \end{tabular}%}
    \caption{Pareto optimal point density \% (ratio of \#extracted optimal points to \#function evaluations). HNPF finds many more optimal points with many fewer function evaluations.}
    \label{tab:eval-pk}
\end{table}

% \textbf{Table \ref{tab:eval-pk}} shows a comparison between the density of point produced by HNPF and PK. We extract point with a much higher density and an even spread.

\section{Verification Case for Fairness} \label{app:rss}

We now demonstrate the nature of the extracted Pareto front under a uniform distribution for the problem setting in \textbf{Section \ref{sec:test}}. While the scenario presented in \citet{gao2019fair} considered points being drawn from a Binomial distribution, it is highly improbable to reach a relevance value of 1. We therefore consider a unbiased uniform distribution, where we sample integers uniformly between $[0,48]\times [0,48]$ for both relevance ($R$) and entropy ($G$). 

\begin{figure}[ht]
    \centering
     \begin{subfigure}{0.4\linewidth}
      \centering
      \includegraphics[width=\linewidth]{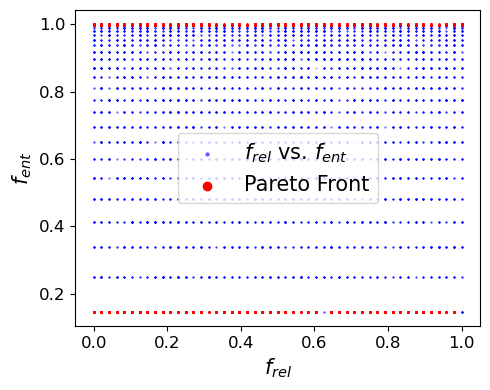}
      \caption{Function Domain}
    \end{subfigure}\qquad
    \begin{subfigure}{0.4\linewidth}
      \centering
      \includegraphics[width=\linewidth]{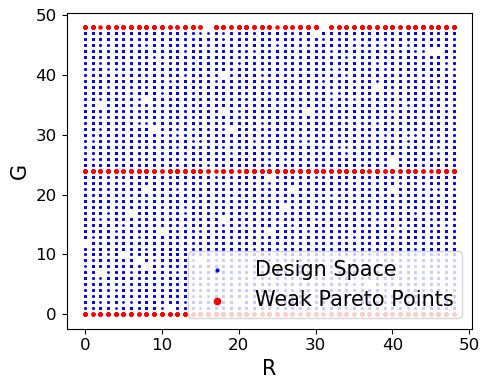}
      \caption{Variable Domain}
    \end{subfigure}
    \caption{Weak Pareto Front for RSS dataset under a uniform setting. Note that now both min-min and the max-max solutions are achievable, as these points are visible to the network.}
    \label{fig:paretorss-unif}
\end{figure}

In \textbf{Fig. \ref{fig:paretorss-unif}} we show the extracted {\em weak} Pareto front from the HNPF neural network. Under the uniform data distribution, the maximum value of achievable entropy corresponds to $G=24$. Similarly, the minimum achievable entropy value corresponds to both ($G=0,48$) which indicates that all documents belong to only one source. The neural network extracts the entire max-max {\em weak} Pareto front which when post-processed using the Pareto filter, results in the expected $(1,1)$ solution in the function domain. Note that the Stage-1 of HNPF is capable of extracting the entire {\em weak} Pareto front for any MOO problem. The middle red line in the variable domain corresponds to the top red line in the function domain. The two remaining red lines (top and bottom) in the variable domain collapses to the bottom in the function domain showing the weak Pareto front for a min-min MOO problem. Applying the Stage-2 Pareto filter now gives $(0,0)$ as the optimal min-min solution. 

\end{document}